\documentclass[12pt]{article}

\usepackage{placeins}      % ------- ADD BY RONAN -------
\usepackage{scicite}
\usepackage[font={sf}]{caption}
\usepackage{times}
\usepackage{subcaption}
\usepackage[table,xcdraw]{xcolor}
\usepackage{booktabs}
\usepackage{siunitx}
\usepackage{todonotes}
\usepackage{ulem}
\usepackage{graphicx}
\usepackage{amsmath,amssymb,amsfonts}
\graphicspath{ {figures/} }
\usepackage{parskip}
\usepackage[misc]{ifsym}
\usepackage[pdfborder={0 0 0},pdftex]{hyperref}

\usepackage{xcolor, soul}
\sethlcolor{green}
\renewcommand{\hl}[1]{#1}

\newenvironment{sciabstract}{%
\begin{quote} \bf}
{\end{quote}}

\title{\vspace{-2.0cm}Low Voltage Electrohydraulic Actuators for Untethered Robotics}

\author
{Stephan-Daniel Gravert$^{1}\dagger$, Elia Varini$^{1}\dagger$, Amirhossein Kazemipour$^{1}$, \\Mike Y. Michelis$^{1}$, Thomas Buchner$^{1}$, Ronan Hinchet$^{2}$, \\Robert K. Katzschmann$^{1*}$\\
\\
\normalsize{$^{1}$ Soft Robotics Lab, D-MAVT, ETH Zurich, Switzerland}\\
\normalsize{$^{2}$ Computational Robotics Lab, D-INFK, ETH Zurich, Switzerland}\\
\\
\normalsize{$\dagger$ These authors contributed equally to this work.}
\\
\normalsize{$^*$ Corresponding author: \href{mailto:rkk@ethz.ch}{rkk@ethz.ch}}
}
\date{}
\begin{document} 
\captionsetup[figure]{labelfont={bf},name={Fig.},labelsep=period}
\captionsetup[table]{labelfont={bf},name={Tab.},labelsep=period}
%\baselineskip24pt  --------> RONAN: DISABLE DOUBLE LINE SPACING to ease writing <---------

\maketitle 

\begin{sciabstract}
Rigid robots can be precise in repetitive tasks, but struggle in unstructured environments. 
Nature's versatility in such environments inspires researchers to develop biomimetic robots that incorporate compliant and contracting artificial muscles. 
Among the recently proposed artificial muscle technologies, electrohydraulic actuators are promising since they offer performance comparable to that of mammalian muscles in terms of speed and power density. 
However, they require high driving voltages and have safety concerns due to exposed electrodes. 
These high voltages lead to either bulky or inefficient driving electronics that make untethered, high-degree-of-freedom bio-inspired robots difficult to realize.
Here, we present hydraulically amplified low voltage electrostatic (\hl{HALVE}) actuators that match mammalian skeletal muscles in average power density (50.5\,W\,kg$^{-1}$) and peak strain rate (971\,\%\,s$^{-1}$) at a driving voltage of just 1100\,V.
This driving voltage is approx. 5-7 times lower compared to other electrohydraulic actuators using paraelectric dielectrics. Furthermore, \hl{HALVE actuators} are safe to touch, waterproof, and self-clearing, which makes them easy to implement in wearables and robotics. We characterize, model, and physically validate key performance metrics of the actuator and compare its performance to state-of-the-art electrohydraulic designs. Finally, we demonstrate the utility of our actuators on two muscle-based electrohydraulic robots: an untethered soft robotic swimmer and a robotic gripper. We foresee that \hl{HALVE actuators} can become a key building block for future highly-biomimetic untethered robots and wearables with many independent artificial muscles such as biomimetic hands, faces, or exoskeletons.
\end{sciabstract}

\subsection*{Summary}
Reducing voltages in electrohydraulic actuators to enable easier implementation in untethered biomimetic robots.

\section*{Introduction}
\subsection*{Problem addressed}
    Rigid robots excel at precision and repetitive tasks but typically fall short in unstructured terrain. Therefore, researchers try to mimic natural organisms, which are well adapted to operate effectively in unstructured environments~\cite{Hawkes2021Hard}. Nature's musculoskeletal architecture inspires roboticists to widen their design space and create artificial muscles that mimic their natural counterparts for integration into biomimetic robots and wearables.

    Electromagnetic actuators have been used to mimic musculoskeletal designs by placing them with tendons outside of joints \textit{e.g.}, Kengoro humanoid with 114 degrees of freedom~\cite{asano_human_2016} or robotic hands~\cite{tuffield_shadow_2003,odhner_compliant_2014,garcia-garcia_tactilegcn_2019}.
    While electromagnetically actuated tendon-driven robots show many degrees of freedom, they lack the desired dexterity, adaptability, softness, and efficiency in natural musculoskeletal organisms.
    %Soft fluidic actuators
    Soft fluidic actuators are inherently more compliant and require less energy in static load scenarios than electromagnetic actuators~\cite{seok_design_2013}. For example, McKibben Pneumatic Artificial Muscles (PAM) have high force density and can achieve strain values between \qtyrange{40}{300}{\percent}~\cite{hawkes_design_2016}, which is more than skeletal muscles'~\qtyrange{20}{40}{\percent} strain~\cite{Xingrui2020HighStrainPeano}. However, soft fluidic actuators require thick pneumatic supply lines, pressure valves, and bulky compressors to operate~\cite{hitzmann_anthropomorphic_2018}.

    Electrostatic actuators directly convert electrical energy into mechanical energy through electromechanical coupling within the muscle~\cite{yasa_overview_2023}. They are thus an order of magnitude more energy efficient (\qty{\sim 20}{\percent}~\cite{Rothemund2021HASEL}) than soft fluidic actuators (\qty{\sim 3}{\percent}~\cite{chun_towards_2019}). They can also be inherently compliant and offer self-sensing capabilities~\cite{kellaris2018peano}. Due to these beneficial properties, many muscle-like electrostatic actuators have been proposed~\cite{acome2018hydraulically,kellaris2018peano,kellaris2021spider,mitchell2019toolkit,Taghavi2018ElectroRibbonActuators,Sirbu2021ElectrostaticBellowMuscle,Duduta2019PotenzialDEA}, of which hydraulically amplified self-healing electrostatic actuators (HASELs) have emerged as a promising candidate by combining the versatility of fluidic actuators (PAMs) with electrostatic principles. Various HASEL designs exhibit a maximum strain of \qtyrange{15}{24}{\percent}, a blocking force of \qtyrange{18}{45}{\newton}~\cite{kellaris2019analytical,Xingrui2020HighStrainPeano}, and a peak specific power comparable to that of mammalian muscle~\cite{kellaris2018peano}. Manufacturing of HASELs is easier than for dielectric elastomer actuators (DEAs)~\cite{acome2018hydraulically}, they can be made from biodegradable materials~\cite{Rumley2023BioHASEL} and they don't require prestreching or rigid frames to achieve high contracting strain values~\cite{kellaris2018peano}.
    
    HASELs show promising performance metrics but are currently difficult to implement into complex untethered soft machines with an acceptable power-to-weight ratio because they require high driving voltages (\qty{> 6}{\kilo\volt}) for operation, limiting robots to bulky power supplies. Additionally, exposed electrodes limit their use in proximity to humans and lead to safety concerns. Although the initial designs of HASEL were self-healing, the latest designs made from thermoplastics do not show self-healing properties~\cite{kellaris2018peano}.

\subsection*{Paper's importance}
    To develop autonomous robots that can mimic musculoskeletal organisms, a high number of independent actuation channels is required. We aim to develop a suitable, compliant artificial muscle system that includes both the actuator and its driving electronics in an untethered small form factor while being safe, fast, powerful, and efficient. Decreasing the operating voltage of HASELs is a key step to enable high-performance untethered soft robots. Our approach to reducing voltages has the potential to be applied to other similar electrohydraulic actuators~\cite{Helps2022LiquidAmplifiedZippingActuators,Garrad2022DielectrophoreticRatchetingActuator,Han2020HapticDeepThing,Hartmann2021Lense,Kim2021Wheel,Schlatter2020PrintingActuators,Taghavi2018ElectroRibbonActuators,Sirbu2021ElectrostaticBellowMuscle}.

\subsection*{Background and related work}
    Designing electrostatic muscles that operate around a few hundred volts is a common strategy to reduce the size, complexity, and cost of the driving electronics~\cite{mitchell2022pocket}. For example, an mm-sized robotic insect was developed by reducing the required driving voltages of its DEA muscles to below \qty{500}{\volt} to allow for tiny power electronics (millimeter-scale)~\cite{ji2019autonomous}. A plethora of low-cost electronic components is available operating in the 100s of volts to serve the established field of piezoelectric devices~\cite{mitchell2022pocket}. The footprint and price of commercially available metal-oxide-semiconductor field-effect transistors (MOSFETs) decrease by an order of magnitude when transitioning from devices operating in the 1000s of volts to those operating in the 100s of volts~(see Fig.~\ref{figS:storyStuff} for an overview). The required actuation voltage of HASELs can be decreased by reducing the thickness of the dielectric material or by increasing its permittivity ~\cite{kirkman2021electromechanics,kellaris2019analytical,moretti2018analysis,rothemund2019inhomogeneous}. Flexible high relative permittivity polymers have been successfully used to achieve high Maxwell stresses at low voltages (\qty{< 500}{\volt}) in electrostatic devices used for haptic~\cite{hinchet2020high,Hinchet2022clutch,Leroy2020HAXEL,Leroy2023HAXEL}. 
    \hl{A patent proposed HASELs with a multilayer structure where different layers can independently  be selected for their electrical and physical properties}~\cite{patent_keplinger}. \hl{Increased load strain capabilities were tested for a PVDF copolymer (PVDF-HFP) at a high voltage of} \qty{5}{\kilo\volt}~\cite{patent_keplinger}.
    
    Recently Mitchell et al.~\cite{mitchell2022pocket} developed a smartphone-sized~(\qty{223}{\cubic\cm}, \qty{250}{\gram}) 10-channel \qty{10}{\kilo\volt} unipolar power supply. They use high voltage optocouplers, which allowed them to reduce the power supply in size, but at the cost of efficiency (estimated 7\% at low load and 37\% at max load) and high price (\$1200). Also, driving circuits based on high voltage optocouplers consume high amounts of energy during idling to ground the actuators (approx. \qty{0.25}{\watt} per actuator), which makes them unpractical for untethered robots. In contrast, MOSFETs, such as Infineon's IPN95R3K7P7m, offer a more attractive alternative. These high voltage MOSFETs provide a significantly smaller footprint (one order of magnitude smaller), a lower price (two orders of magnitude cheaper), and a dramatically decreased power consumption (two orders of magnitude lower)~(Tab.~\ref{tabS:switchingComponent}). The rapid switching speed of MOSFETs (on the order of nanoseconds) compared to optocouplers (on the order of microseconds) and their higher output current (\qty{2}{\ampere} versus \qty{0.5}{\milli\ampere}) hold promise for advancing real-time control in HASELs. However, despite these advantages, MOSFETs have not yet been adopted for HASELs due to their restricted switching voltage, limited to voltages below \qty{5}{\kilo\volt}.

\subsection*{Contributions}
    This paper contributes to the field of robotics a muscle-like, contracting actuator that achieves an average power density (\qty{50.5}{\watt\per\kilo\gram}) and peak strain rate (\qty{971}{\percent\per\second}) comparable to mammalian skeletal muscle (typical \qty{50}{\watt\per\kilo\gram}, peak \qty{500}{\percent\per\second})~\cite{Madden2004MuscleHuman,Hunter1992MaxStrainRate,Mirfakhrai2007StrainRate,Rothemund2021HASEL} at an actuation voltage of just \qty{1100}{\volt}~(Fig.~\ref{fig_introduction}B). This driving voltage is 5-7 times lower compared to other electrohydraulic actuators using paraelectric dielectrics. Additionally, \hl{HALVE actuators} are safe to touch, waterproof, and self-clearing making them easy to implement in robotics or wearables. The self-clearing property enables the actuator to endure multiple dielectric breakdowns during operation without a significant decrease in performance.
    %At this voltage the actuator also achieves a free strain of 9\%.
    \hl{Using a multilayer dielectric structure for a Peano-HASEL}~\cite{patent_keplinger} \hl{consisting of a force-bearing and dielectric layer allowed us to independently optimize the actuator's electrical and mechanical properties while shielding the electrode} (see Fig.~\ref{fig_introduction}A). We used high permittivity thin polymer layers of \hl{PVDF terpolymer} (PVDF-TrFE-CTFE) to reduce driving voltages. To accurately predict the force output of our actuators, we modeled and physically validated the electrostatic force with the effective dielectric constant~\cite{Chu2006} of the ferroelectric material. We provide extensive experimental characterizations of the \hl{HALVE actuators} and comparisons with state-of-the-art Peano-HASEL actuator designs for key performance metrics. Finally, to demonstrate how these actuators enable untethered robotic applications, we integrate \hl{HALVE actuators} into a functional muscle system that includes a small, custom-made, and efficient power supply, creating a gripper and a robotic fish as demonstrators.

\section*{Results}
    \subsection*{Actuator overview}
        %\paragraph{HASEL structure}
        In HASELs, a thin thermoplastic film, typically either biaxially-oriented polypropylene (BOPP) or biaxially-oriented polyethylene terephthalate (BoPET) (Fig.~\ref{fig:HASEL_Performance}B), serves two functions: as the actuator's structural element and as the dielectric layer for the electrostatic actuation. This duality of function constrains the material choice, as the material cannot be optimized for high tensile strength and dielectric properties independently. Electrohydraulic actuators can be further optimized by decoupling these two functions into separate elements of the actuator structure, allowing for more specific material selection.
        
\begin{figure}[h]
    \centering
    \includegraphics[width=\textwidth]{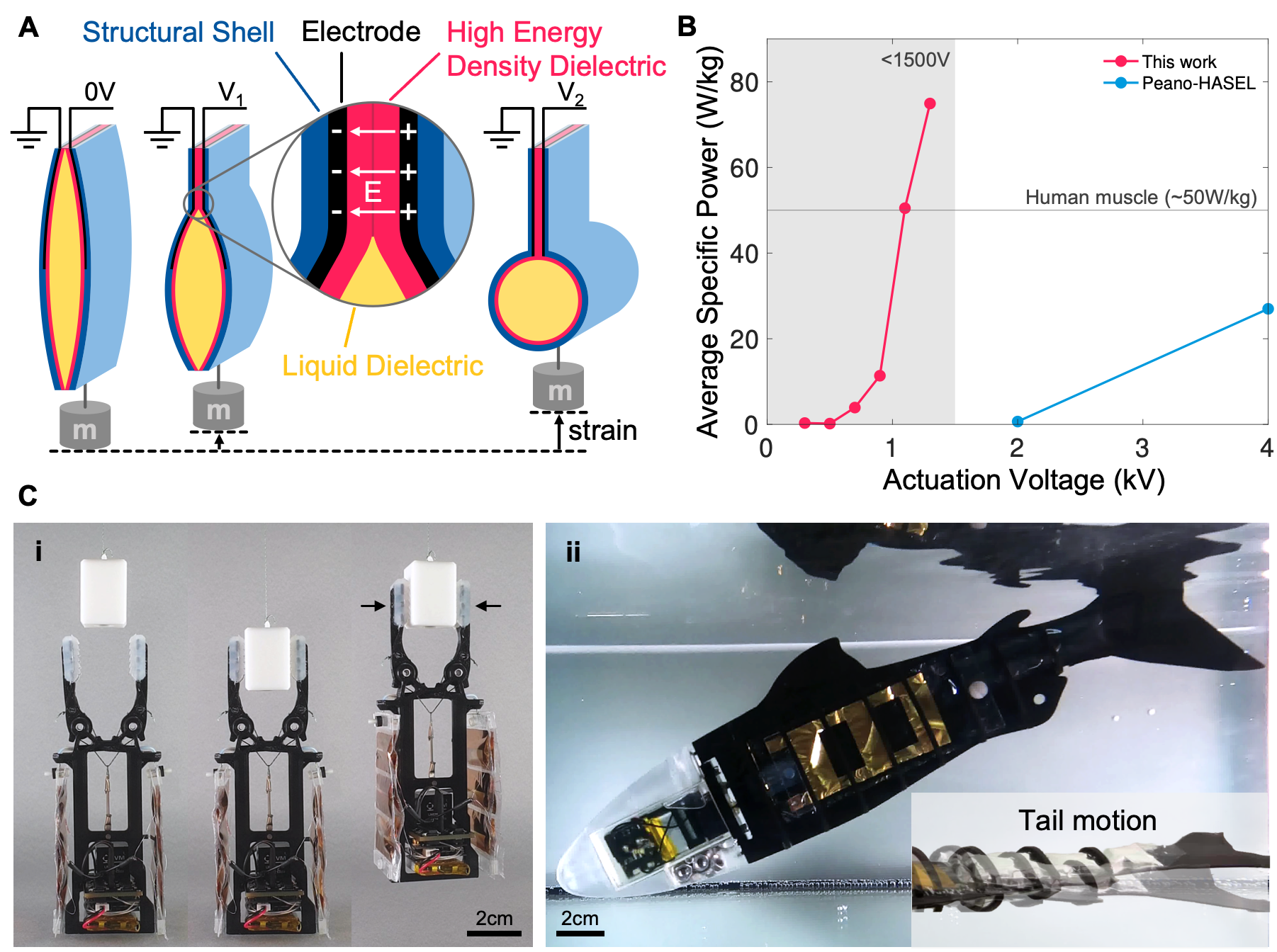}
    \caption{\textbf{A low voltage muscle system for untethered electrostatic robots.} (\textbf{A})~General structure of a \hl{HALVE actuator} and working mechanism. The electrodes zipp-in upon application of voltage potential and displace the dielectric liquid into a progressively more cylindrical shape. The transition from relaxed to fully zipped can be seen with $0V<V_1<V_2$. The structural shell carries the load on the actuator and insulates the electrodes. (\textbf{B})~Average specific power of a \hl{HALVE actuator} (\qty{5}{\micro\meter} PVDF-TrFE-CTFE) versus a Peano-HASEL (\qty{15}{\micro\meter} BoPET) as the dielectric shell with a \qty{300}{\gram} load. The horizontal line indicates the typical specific power of mammalian skeletal muscle at \qty{50}{\watt\per\kilo\gram}~\cite{Madden2004MuscleHuman}. (\textbf{C})~(\textbf{i})~Untethered gripper that can be lifted from the floor. (\textbf{ii})~Fully integrated demonstrator artificial fish floating in the water. The artificial fish measures approx. \qty{28}{\centi\meter} in length.}
    \label{fig_introduction}
\end{figure}

        %\paragraph{OUR device}
        We \hl{use} a different device structure (Fig.~\ref{fig_introduction}A) composed of three elements: First, an outer shell made of a mechanically strong polymer that serves as the structural element and as an outer electric insulation layer that covers the whole device to avoid electrical discharge with the environment. The second element is an electrode that covers a portion of the pouch. The third element is a thin layer of P(VDF-TrFE-CTFE) that serves as a high energy density dielectric layer to boost electrostatic actuation performance and lower the actuation voltage. Finally, the central cavity is filled with dielectric oil that serves as a hydraulic amplification medium for the actuator.
        
        %\paragraph{working mechanism}
        \hl{We use the Peano-HASEL geometry}~\cite{kellaris2018peano} \hl{to study the capabilities of the three-layer material composite structure}~\cite{patent_keplinger} \hl{at low voltages below} \qty{1.1}{\kilo\volt}, terming the resulting actuator \hl{Hydraulically Amplified Low Voltage Electrostatic (HALVE) actuator}. The principle of operation is the following: When a voltage difference is applied between the electrodes, opposite electric charges build up in each electrode. The resulting Coulomb force attracts the electrodes that squeeze the dielectric oil away from the electrodes to the bottom part of the pouch, which deforms progressively into a cylindrical shape. This change in geometry shortens the length of the actuator, thus generating mechanical work (Fig. \ref{fig_introduction}A).

    \subsection*{Model}
\begin{figure}[h]
    \centering
    \includegraphics[width=0.85\textwidth]{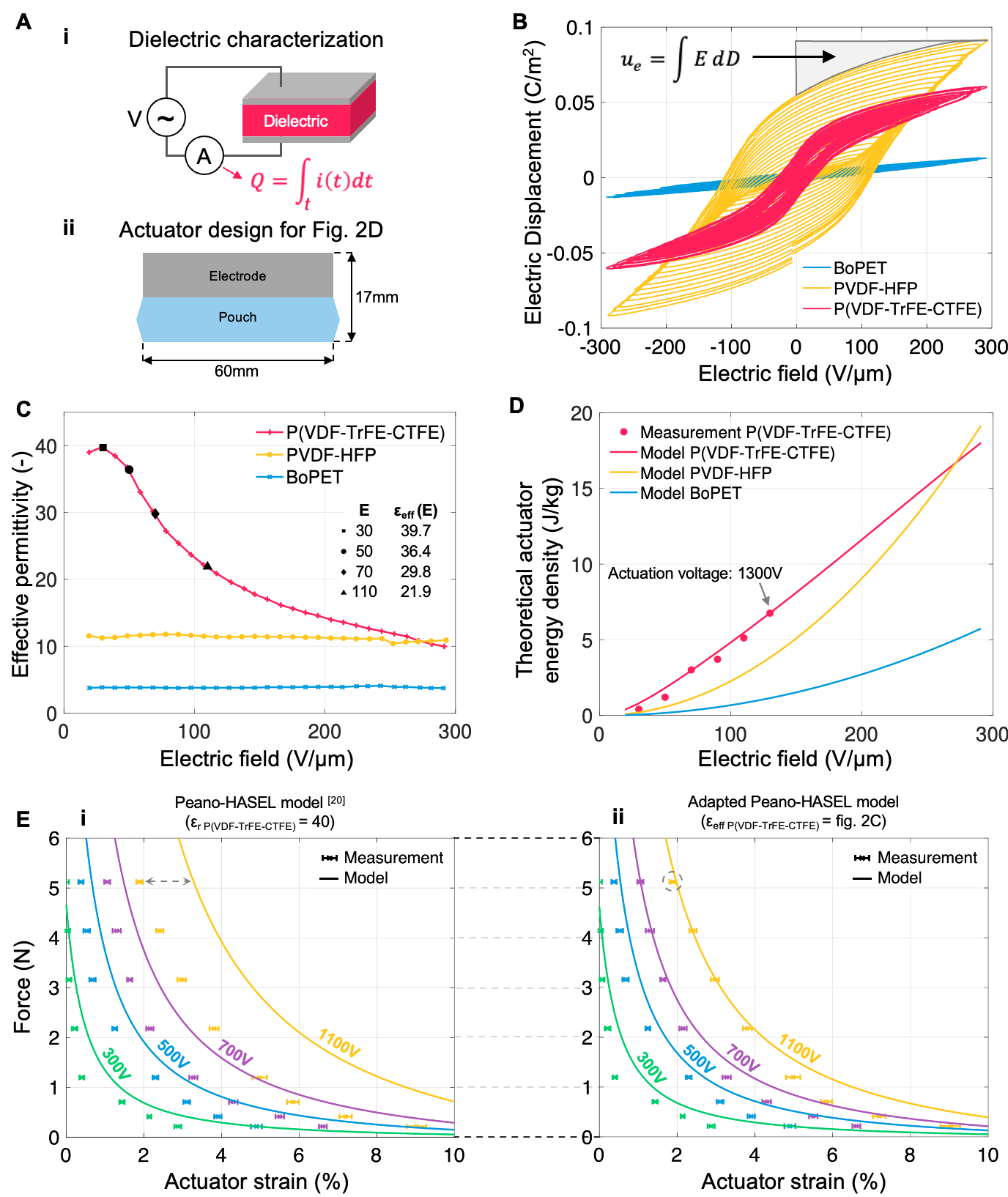}
    \caption{\textbf{Effective dielectric constant for model predictions of \hl{HALVE actuators} force/strain behavior and model validation.} (\textbf{A})~(\textbf{i})~Depiction of the dielectric measurement setup used to measure the data points for Fig. \ref{fig:hasel_model}B. (\textbf{ii})~Actuator design specifications for which the actuator energy densities are calculated in \ref{fig:hasel_model}D. (\textbf{B})~Discharge D-E curves for BoPET, PVDF-HFP, and P(VDF-TrFE-CTFE), the highlighted gray area corresponds to the energy density integral of PVDF-HFP at \qty{300}{\volt\per\micro\meter}. (\textbf{C})~Effective permittivities calculated using Eq. \ref{eq.effper} using the experimental data shown in \ref{fig:hasel_model}B. (\textbf{D})~Prediction for the resulting actuator energy densities for different dielectrics. (\textbf{E})~Force/strain measurements and model prediction. (\textbf{i})~Model is plotted with a constant dielectric constant of 40 as suggested in \cite{kellaris2019analytical}. (\textbf{ii})~Model is plotted with effective dielectric constant values from Fig. \ref{fig:hasel_model}C.}
    \label{fig:hasel_model}
\end{figure}

        Kellaris et al.~\cite{kellaris2019analytical} have proposed a quasi-static model of Peano-HASELs based on the total free energy $U_t$: 
        \begin{equation}    U_t=Fx-\frac{1}{2} CV^2     
        \label{eq:freeenergy}
        \end{equation}
        where $F$ is the mechanical force generated by the actuator over the contraction distance $x$. The product of both represents the mechanical work produced by the actuator. $C$ is the capacitance between the electrodes of the actuator and $V$ is the voltage applied. For a given design, the free energy is constant once it is at equilibrium. To maximize the mechanical work, the electrical energy must be maximized. To achieve this, one can raise the voltage $V$, but it is advantageous to keep it as low as possible for wearable and autonomous applications. Instead, we want to increase the capacitance $C$. Considering the zipped region of a \hl{HALVE actuator} as an ideal parallel plate capacitor with a homogeneous polymer insulating layer and neglecting the oil thickness remaining at the interface, the capacitance can be approached by the formula:
        \begin{equation}
        C=\frac{A \epsilon_0 \epsilon_r}{t}=\frac{wl \epsilon_0 \epsilon_r}{t}\end{equation}
        where $A$ is the overlapping area of the electrodes, $l$ their length, $w$ their width, $t$ is the thickness of the dielectric, $\epsilon_{0}$ the vacuum permittivity and $\epsilon_{r}$ is the relative permittivity constant of the insulator. 
        % \paragraph{High-K materials}
        Selecting a flexible insulator with higher permittivity, such as a Polyvinylidene fluoride co-polymer like P(VDF-HFP) or terpolymer (PVDFter) such as P (VDF-TrFE-CTFE)~\cite{hinchet2020high} enables us to increase the capacitive electrical energy for a given voltage and design. Theoretically, replacing the commonly used 15 $\mu$m thick BoPET film~\cite{kellaris2021spider} ($\epsilon_r=3.3$) with a 5 $\mu$m thick P(VDF-TrFE-CTFE) dielectric layer ($\epsilon_r=40$) could reduce the required operating voltage by approximately 7.5 times. However, it is more complex in practice because the permittivity of such high-permittivity insulators is not constant, but depends on the amplitude and frequency ($f$) of the applied electric field ($E$). Additionally, the fabrication process and material quality affect the insulator's permittivity. Thus, to achieve the same electrical energy, the voltage input can be decreased by a ratio of:
        \begin{equation}    \frac{V_{PET}}{V_{PVDFter}} = \sqrt{\frac{t_{PET} }{\epsilon_{rPET} \ t_{PVDFter}}  \times \epsilon_{eff\ PVDFter}(E,f)}  \end{equation}
        where $\epsilon_{eff}$ is the effective permittivity. Using the generalized mean value theorem for integrals, the effective permittivity can be calculated by measuring the energy density $u_e$ of the dielectric material~(see Text S1. for proof). The energy density $u_e$ of a dielectric is equal to the integral of the discharge D-E curve, which is highlighted in Fig. \ref{fig:hasel_model}B:~\cite{Yao2011, Chu2006}
        \begin{equation}
        u_e = \int EdD = \frac{1}{2}\epsilon_{eff}(E,f)\epsilon_0E^2
        \label{eq.effper}
        \end{equation}
        where E is the electric field and D is the electric displacement. As expected, measurements show that at low frequencies (\qty{2}{\hertz}) the effective permittivity of BoPET and P(VDF-HFP) are almost constant for electric fields up to 300 V/$\mu$m. However, the measured effective permittivity of the P(VDF-TrFE-CTFE) peaked at 39.5 at 30 V/$\mu$m and then decreased to 10 at 300 V/$\mu$m (Fig. \ref{fig:hasel_model}C). Measurements suggested that above 300 V/$\mu$m P(VDF-HFP) has a higher effective dielectric constant than P(VDF-TrFE-CTFE). The dielectric material characterization also suggests that a P(VDF-HFP) based device should be able to outperform a BoPET-based HASEL in force by 3.5 times over the entire voltage range.

        %explain force/strain curve
        Kellaris et al. \cite{kellaris2019analytical} derived the force/strain relationship of the Peano-HASEL from its total free energy (Eq. \ref{eq:freeenergy}). To adapt the model to ferroelectric dielectrics we swapped the relative dielectric constant $\epsilon_r$ with the effective dielectric constant $\epsilon_{eff}(E,f)$ previously introduced:
        \begin{equation}
            F = wt \frac{cos(\alpha)}{1-cos(\alpha)} \epsilon_0 \epsilon_{eff}(E,f)E^2
            \label{eq:ForceStrainPeanoHASEL}
        \end{equation}
        where $\alpha$ is the opening angle, $w$ is the width of the Peano-HASEL and $t$ is the thickness of the dielectric as defined in \cite{kellaris2019analytical}. For a \hl{HALVE} device, $t$ corresponds to the thickness of the high energy density dielectric layer between the electrode and the oil. The adapted force / strain equation allowed us to predict actuator energy densities by integrating the area under the modeled force/strain curve divided by the weight of the actuator~\cite{Duduta2019PotenzialDEA}. We used Eq. \ref{eq:ForceStrainPeanoHASEL} and the values from Fig. \ref{fig:hasel_model}C to calculate the theoretical actuator energy densities of \hl{HALVE} devices. Fig. \ref{fig:hasel_model}D shows the predicted actuator energy densities of an actuator made of P(VDF-TrFE-CTFE), PVDF-HFP, or BoPET as dielectric using Eq. \ref{eq:ForceStrainPeanoHASEL} and the values from Fig. \ref{fig:hasel_model}C. We can see that for P(VDF-TrFE-CTFE) the relationship between actuator energy density (force of the actuator) and the applied electric field is linear, whereas it is quadratic for PVDF-HFP and BoPET. This linear force response of PVDF terpolymers to applied voltages has been previously observed in literature~\cite{Chu2006, Nishimura2022Lowering}. We observed that an actuator made from PVDF-HFP achieves a higher actuator energy density above \qty{280}{\volt\per\micro\meter}, due to the effective permittivity of P(VDF-TrFE-CTFE) dropping below the value of PVDF-HFP. Fig. \ref{fig:hasel_model}E shows a high discrepancy at a high voltage between the theoretical force-strain curves of \hl{HALVE} devices calculated considering a constant permittivity of 40 on the left (i) and using the effective permittivity of P(VDF-TrFE-CTFE) on the right (ii). 
        
    \subsection*{Performance characterization}
\begin{figure}[h]
    \centering
    \includegraphics[width=0.85\textwidth]{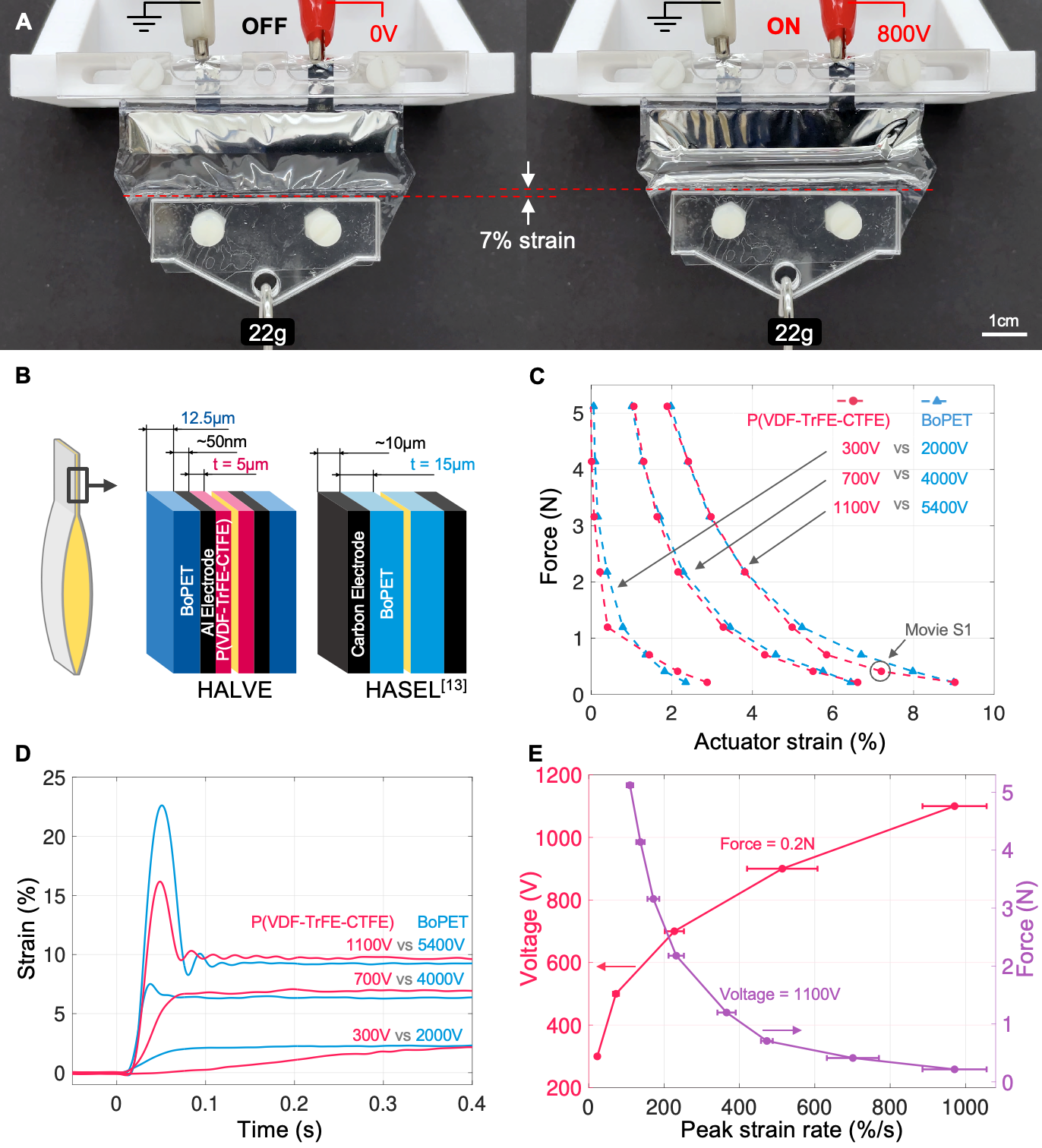}
    \caption{\textbf{Performance characterization of the \hl{HALVE actuator}.}  (\textbf{A})~\hl{HALVE actuator} at rest and contracted during actuation. (\textbf{B})~Strain-force curves of a \qty{5}{\micro\meter} P(VDF-TrFE-CTFE) actuator and a \qty{15}{\micro\meter} BoPET actuator, in pairs of actuation voltages which lead to the same theoretical Maxwell-stress term. 
    (\textbf{C})~Specific work plotted against actuation voltage of a \hl{HALVE actuator} with \qty{5}{\micro\meter} P(VDF-TrFE-CTFE) as solid dielectric layer and traditional HASEL actuator with \qty{15}{\micro\meter} BoPET as solid dielectric layer.
    (\textbf{D})~Comparison between strain responses to voltage step inputs of different voltage amplitudes of a P(VDF-TrFE-CTFE) \hl{HALVE actuator} while lifting a \qty{22}{\gram} weight.
    (\textbf{E})~Left axis: relation between actuation strain rate and actuation voltage of a P(VDF-TrFE-CTFE) \hl{HALVE actuator} at a constant force of \qty{0.22}{\newton}. Right axis: relation between actuation strain rate and force at a constant voltage of \qty{1100}{\volt}.}
    \label{fig:HASEL_Performance}
\end{figure}

        %Intro
        To validate the model, the quasi-static relation between actuation force and strain was experimentally characterized by measuring the displacement produced by the actuators (Fig.~\ref{fig:HASEL_Performance}A) at a given actuation voltage and with a known weight attached to it (Fig. \ref{Sfig:measurement}A). \hl{HALVE} devices were manufactured from BoPET film as structural shell and with a \qty{5}{\micro\meter} thick P(VDF-TrFE-CTFE) layer as solid dielectric~(Fig. \ref{fig:HASEL_Performance}B). To compare actuator performance, the same electrode and pouch geometry was used in all characterizations (Fig. \ref{fig:hasel_model}Aii). 

        %Fig. 2E
        Due to its lower permittivity at high electric fields, the model predicted that the force of a \hl{HALVE device} actuated at \qty{1100}{\volt} reaches approx. \qty{5}{\newton} at only 2 $\%$ strain instead of 4 $\%$. This prediction is in agreement with the measured values (Fig. \ref{fig:hasel_model}E). Still, both models overpredicted the performance at high strain, which could be due to imprecise filling amounts and edge constraints. As expected, both models predicted similar performance at low electric fields, as the relative and effective dielectric constants have similar values.

        %Comparison to PET-HASEL
        To gain a more comprehensive understanding of the strengths and limitations of our approach, we compared \hl{HALVE actuators} to Peano-HASEL actuators.  Peano-HASELs actuators were manufactured from thermoplastic polymers as suggested in literature~\cite{kellaris2018peano,mitchell2019toolkit,kellaris2021spider}, with \qty{15}{\micro\meter} thick BoPET film as the solid dielectric layer with carbon ink electrodes (Fig. \ref{fig:HASEL_Performance}B). Both HASEL and \hl{HALVE} devices had the same size and geometry (Fig.~\ref{fig:hasel_model}Aii).
        
        %Describes Figure 3C
        To mitigate any charge retention in the dielectrics we applied a voltage signal with an alternating polarity (bipolar) to both actuator types~\cite{hinchet2020high, rothemund2020dynamics}. We recorded the strain curves at various voltage amplitudes and weights for both HASEL and \hl{HALVE} devices. As expected, the force decreased with increasing strain (Fig. \ref{fig:HASEL_Performance}C), following a profile in agreement with literature~\cite{kellaris2019analytical} and with commercially available Peano-HASELs~\cite{artimus_datasheet}. As can be seen in Fig.~\ref{fig:HASEL_Performance}C \hl{HALVE actuators} had comparable quasi-static performance to HASELs at a voltage between 4.9 and 6.6 times lower. Movie~S1 shows a \hl{HALVE} device that was actuated at \qty{1100}{\volt} and produced approx. 7\% strain lifting a \qty{42}{\gram} weight. The P(VDF-TrFE-CTFE) layer of \hl{HALVE} devices should be able to withstand field strengths up to \qty{350}{\volt\per\micro\meter}~\cite{hinchet2020high}, but in practice, breakdowns occurred regularly above \qty{120}{\micro\meter}, potentially due to defects and impurities introduced during manufacturing. Peano-HASEL actuators made from industrial grade BoPET films were able to tolerate higher field strengths (\textgreater \qty{300}{\volt\per\micro\meter}) and therefore were able to outperform \hl{HALVE actuators'} strain at voltages above \qty{6}{\kilo\volt}.

        %Measured actuator energy density
        To further validate the adapted model we compared the model prediction for the actuator energy density with values derived from measured force/strain curves (Fig.~\ref{fig:hasel_model}D). To determine the actuators' energy density across the actuation range, we calculated the surface below the measured force-strain curve~\cite{Duduta2019PotenzialDEA}. For the integration of measured force-strain curves, we performed an optimized curve-fit onto the force-strain measurement data with a box-constrained optimization/system identification on 8 relevant parameters using the analytical formula derived by Kellaris et al.~\cite{kellaris2019analytical}. The optimization is explained in more detail in the supplementary materials. We observed good matching between the model prediction and the measured values for the actuator energy density (Fig.~\ref{fig:hasel_model}D).
        
        %Dynamic response Fig. 3D & E
        Actuation voltages also affect the dynamic response of HASEL actuators as reported in literature~\cite{kellaris2018peano,rothemund2020dynamics}. As shown in Fig.~\ref{fig:HASEL_Performance}D, the higher the voltage applied to the actuators, the faster the actuation for the same force.  For Peano-HASELs, we observed that at \qty{2000}{\volt}, the strain slowly increases until reaching the stationary regime. At higher voltages Peano-HASELs show under-damped oscillations and strain overshoot as reported by \cite{kellaris2018peano}. We observed the same phenomena with \hl{HALVE} devices, but with more damping because the voltages involved are lower. Similar to Peano-HASELs~\cite{kellaris2018peano} the strain rate (Fig. \ref{fig:HASEL_Performance}E) is proportional to the square of the voltage, and reversely proportional to the actuation force (Fig. \ref{fig:HASEL_Performance}E). The strain rate for \hl{HALVE actuators} peaked at \qty{971}{\percent\per\second} at \qty{1100}{\volt}, lifting a \qty{22}{\gram} weight, which is comparable to the max. strain rate achieved for BOPP Peano-HASELs~\cite{kellaris2018peano} and about twice the max strain rate in mammalian skeletal muscle (\qty{500}{\percent\per\second})~\cite{Hunter1992MaxStrainRate,kellaris2018peano}. Other parameters can influence the strain rate: analytical models~\cite{rothemund2020dynamics} predict that the HASEL strain rate depends on the actuator geometry and dielectric liquid viscosity. Those parameters could be optimized to improve the strain rate of \hl{HALVE actuators}.

        %Average specific power
        Average and peak specific power were calculated during contraction cycles following the same steps as outlined by Kellaris et al.~\cite{kellaris2018peano} (see Fig.~\ref{figS:specificPowerExplanation} and supplementary information). The comparison of the average specific power of a \hl{HALVE} and Peano-HASEL device is shown in Fig.~\ref{fig_introduction}B for a load of \qty{300}{\gram}. The \hl{HALVE actuator} reaches an average specific power of \qty{50.5}{\watt\per\kilo\gram} at an actuation voltage of \qty{1100}{\volt} and \qty{75}{\watt\per\kilo\gram} at an actuation voltage of \qty{1300}{\volt}. Fig.~\ref{figS:specificPowerGraph} shows the peak and average specific power values for various loads and voltages. The peak specific power in \hl{HALVE} devices reached \qty{132.7}{\watt\per\kilo\gram} at \qty{1300}{\volt} for a load of \qty{300}{\gram} which is comparable to mammalian skeletal muscle, which falls between \qty{50}{\watt\per\kilo\gram} (typical) and \qty{284}{\watt\per\kilo\gram} (maximum)~\cite{Madden2004MuscleHuman}. 
        While our load vs. specific power measurements (Fig.~\ref{figS:specificPowerGraph}) follow the same trajectory reported by Kellaris et al.\cite{kellaris2018peano}, we observe that our actuators achieve the highest specific power at a load of \qty{300}{\gram}, which is higher than the corresponding result (\qty{100}{\gram}) for BOPP Peano-HASELs~\cite{kellaris2018peano}.

    \subsection*{System integration properties}
        \subsubsection*{Interaction with the environment}
            The outer structural shell of the \hl{HALVE actuator} (Fig. \ref{fig_introduction}A) insulates the electrodes and allows the actuator to come into contact with the environment as well as protects the solid dielectric from humidity. Thus, the actuator can be touched during actuation (Fig. \ref{fig:HASEL_Integration}A) and works underwater (Fig. \ref{fig:HASEL_Integration}B). Movie~S2 shows a one-pouch \hl{HALVE} device being touched on the high voltage electrode side, while actuated at \qty{700}{\volt}. Movie~S3 shows a three-pouch \hl{HALVE actuator} being actuated partially submerged in tap water. When a \hl{HALVE actuator} is touched or operated underwater the structural shell can act as a dielectric layer creating a second capacitance towards the outside of the actuator. This capacitor can interfere with the main capacitance between actuator electrodes but is typically negligible as the structural shell has a smaller permittivity and is thicker compared to the internal dielectric. Discharge is unlikely to happen towards the outside of the actuator because the structural shell has a very high breakdown voltage of (\qty{7500}{\volt}). Other methods have been proposed to waterproof HASELs by coating one electrode with adhesive tape and using water as the ground electrode~\cite{Wang2023Jellyfish}. However, the resulting actuator only functions if the electronics and actuator are fully submerged, they can only use unipolar actuation increasing charge retention and the adhesive tape stiffens the actuator limiting its maximum strain. 
        
        \subsubsection*{Self-Clearing}
\begin{figure}[h]
    \centering
    \includegraphics[width=0.9\textwidth]{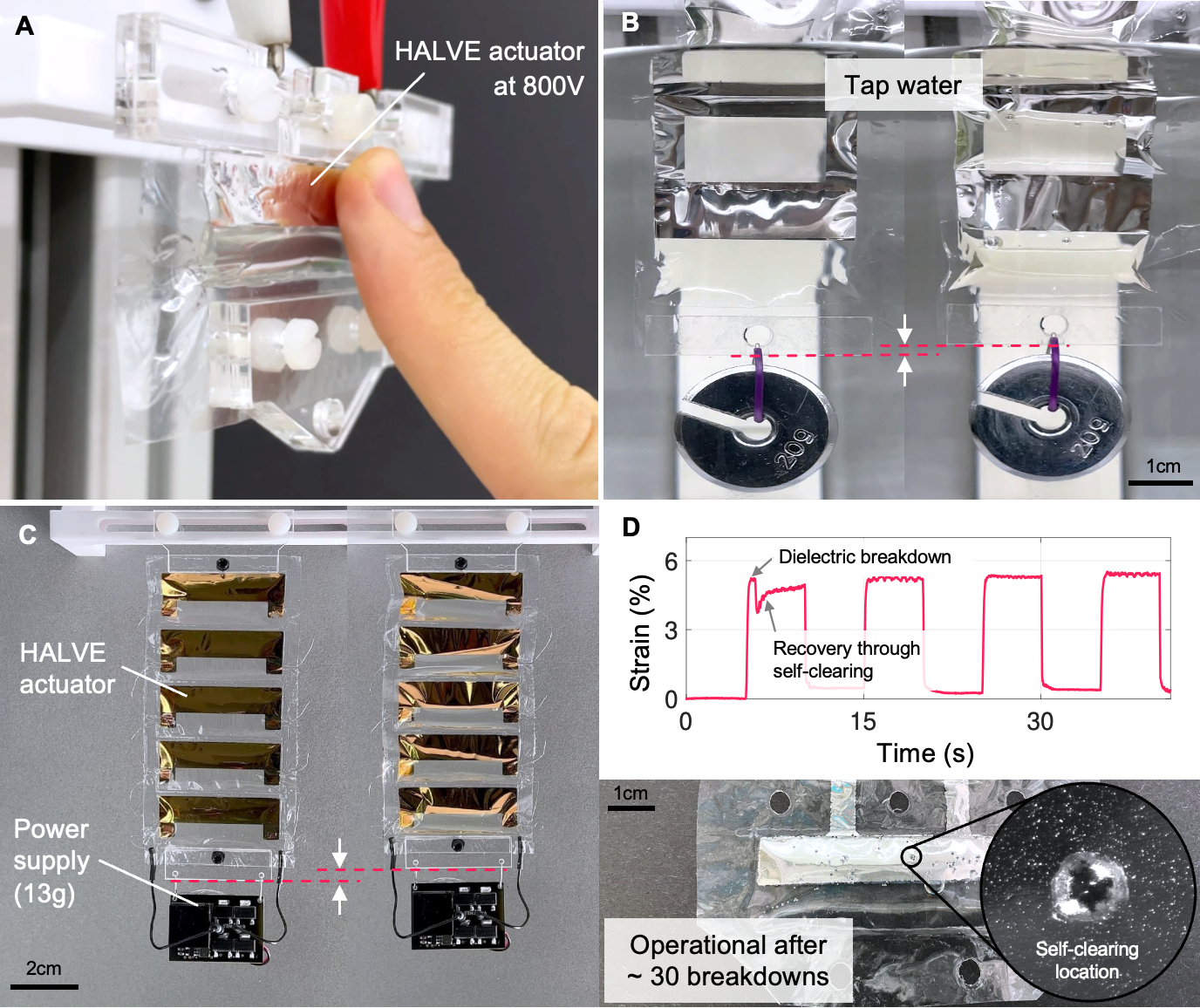}      
    \caption{\textbf{Implementation features of the \hl{HALVE actuator} system.} (\textbf{A})~A \hl{HALVE actuator} is touched on the high voltage side in the zipped state at \qty{800}{\volt}. (\textbf{B})~Three pouch \hl{HALVE actuator} partially submerged in a laboratory beaker filled with tap water lifting a \qty{20}{\gram} weight. (\textbf{C})~Five pouch \hl{HALVE actuator} with chrome/gold electrodes lifting its power supply (\qty{}{\gram}). (\textbf{D})~Self-clearing capabilities of the \hl{HALVE actuator}. (\textbf{i})~Strain curve of a single-pouch \hl{HALVE actuator} lifting a \qty{22}{\gram} weight at a low frequency of \qty{0.1}{\hertz}. The actuator drops slightly in strain after the dielectric breakdown event and recovers its full strain over the next few seconds. (\textbf{ii})~Picture of the same actuator after $\sim 30$ dielectric breakdown events, which is still fully functional.}
    \label{fig:HASEL_Integration}
\end{figure}

            The self-clearing property makes \hl{HALVE} actuators more resilient to manufacturing defects that could create dielectric breakdowns. During a breakdown event, \hl{HALVE actuator's} electrode is immediately destroyed locally, effectively self-clearing the short-circuit~(Fig. \ref{fig:HASEL_Integration}D). Typically, the outer structural shell remains intact, preventing leaks. Thus, the actuators can endure multiple breakdowns and remain operational. Fig. \ref{fig:HASEL_Integration}D shows a \hl{HALVE actuator} that has endured approx. 30 breakdown events while remaining operational. The self-clearing property of the actuators is dependent on voltage and electrode thickness. High voltages or thick electrodes can lead to more violent breakdowns, which can cause damage to the outer structural shell of the actuator. Multiple self-clearing events increase the number of charge carriers in the oil, as burnt particles of the solid dielectric and electrode mix with the oil. This can reduce the performance of that specific pouch. Nonetheless, a multi-pouch \hl{HALVE actuator} will typically survive many breakdown events, whereas a single breakdown event will typically short an entire multi-pouch HASEL actuator. With thermoplastic HASELs, a dielectric breakdown punches a hole into the oil pouch, leaking the oil and typically shorting the electrodes by melting both zipped sides together.

        \subsubsection*{Electronics and system size}
        \hl{HALVE actuator's} lower operation voltage enables small electronics. We developed an autonomous high-voltage power supply, based on the Peta-pico-Voltron design~\cite{schlatter2018peta}, to power the \hl{HALVE} devices, which is shown in Fig.~\ref{figS:electronics}B. Movie S4 shows a five-pouch \hl{HALVE actuator} lifting one of these custom-developed power supplies. As core components, we used a miniature \qty{1}{\kilo\volt} DC/DC converter from HVM Technology and commercially available high voltage MOSFETs (IRF7509TRPBF) ($V_{DS}\leq$\qty{950}{\volt}). The module also includes a microcontroller, radio or Bluetooth antenna, and a \qty{150}{\milli\ampere\hour} LiPo battery and can be extended with other components from the TinyCircuits ecosystem. The entire unit excluding the DC/DC HV converter requires \qty{0.35}{\watt} of power. The DC/DC HV converter requires between \qty{0.05}{\watt} at no load to \qty{0.875}{\watt} at max load. The 2-channel electronic assembly is $42 \times 19 \times 22$ \qty{}{\milli\meter} in size or \qty{11}{\cubic\centi\meter} in volume and weighs \qty{15.5}{\gram} (see Fig.~\ref{figS:electronics}B). The assembly can easily be extended to a higher number of channels by repeating the H-bridge assembly. Each additional channel adds approx. \qty{1.4}{\centi\meter} in length or \qty{1.5}{\cubic\centi\meter} in volume to the power supply. In Fig. \ref{figS:electronics}C we compare the frequency response of a \qty{0.5}{\watt} version with a \qty{0.1}{\watt} version and observe that a more powerful DC/DC converter can improve the frequency response of the actuators. To maintain actuation speeds at similar levels when using a greater number of actuators or those with larger electrode areas (\textit{i.e.}, higher capacitance), a DC/DC converter with higher output power can be employed. The price for the two-channel high voltage power unit excluding the microcontroller, communication module, and battery was $\sim150\$$ for the DC/DC converter plus $\sim35\$$ for all other components. As a comparison a comparable full-bridge \qty{10}{\kilo\volt} 2-channel power supply with high voltage optocouplers as shown by Mitchell et al.~\cite{mitchell2019toolkit} would be $\sim350\$$ for the DC/DC converter and $\sim730\$$ for the required optocouplers at current market price (May 2023).
    
    \subsection*{Untethered robotic demonstrators}
        \subsubsection*{Gripper}
\begin{figure}[h!]
    \centering
    \includegraphics[width=\textwidth]{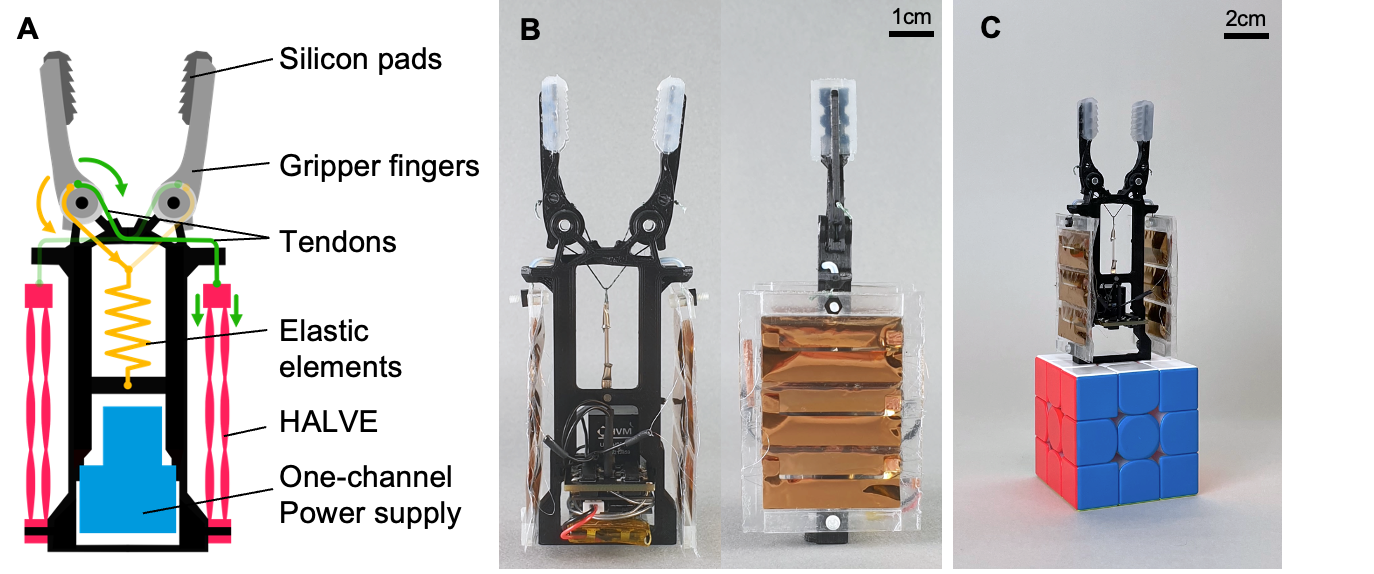}
    \caption{\textbf{Untethered gripper demonstrator powered by \hl{HALVE actuator}.} (\textbf{A})~Schematic view of the untethered gripper. Shown is the tendon system which connects a \hl{HALVE actuator} muscle pack to each gripper finger, as well as the return spring visible in green and yellow respectively. Arrows of the same color mark the direction in which the tendon is pulled and the direction of the resulting torque on the finger's joint. (\textbf{B})~Front, and side views of the untethered gripper. (\textbf{C})~Size comparison of the \hl{HALVE actuator} gripper to a standard-sized Rubiks Cube ($5.6 \times 5.6 \times 5.6$ \qty{}{\centi\meter})}
    \label{fig:Gripper}
\end{figure}

        To demonstrate the potential of \hl{HALVE actuators} with their small form factor control electronics, 
        we developed an untethered gripper powered by \hl{HALVE actuators} (Fig.~\ref{fig:Gripper}). Movie S5 shows the untethered gripper grasping a block made from PLA. The gripper is composed of a main body to which two fingers are attached via revolute joints. 
        The fingertips are equipped with cast silicon pads, to improve the grip. Two packs of \hl{HALVE} muscles are used to generate gripping force. Each pack is composed of two individual actuators connected in parallel and composed of 3 pouches in series. The actuator packs are fixed to the main body at one extremity, while the other is linked to one of the gripping fingers via a tendon. 
        An elastic element, connected to the fingers via another set of tendons, serves as an antagonistic element to the actuators, restoring the gripper to an open state after gripping. In order to power the whole gripper, a one-channel power supply was used (also shown in Fig.~\ref{fig:HASEL_Integration}C). The power supply was powered with a \qty{150}{\milli\ampere\hour} Lithium polymer battery. The combined weight of the battery and power supply was \qty{15}{\gram}. The entire gripper assembly weighs \qty{30}{\gram}, and can fit into a \qty{112}{\milli\meter} (height) by \qty{40}{\milli\meter} (width) by \qty{45}{\milli\meter} (depth) envelope. Fig.~\ref{fig_introduction}C shows how this untethered gripper is able to grasp a smooth plastic (PLA) object with a pinch grasp firmly enough to be lifted into the air.
        
        \subsubsection*{Bio-inspired swimmer}
\begin{figure}[h!]
    \centering
    \includegraphics[width=\textwidth]{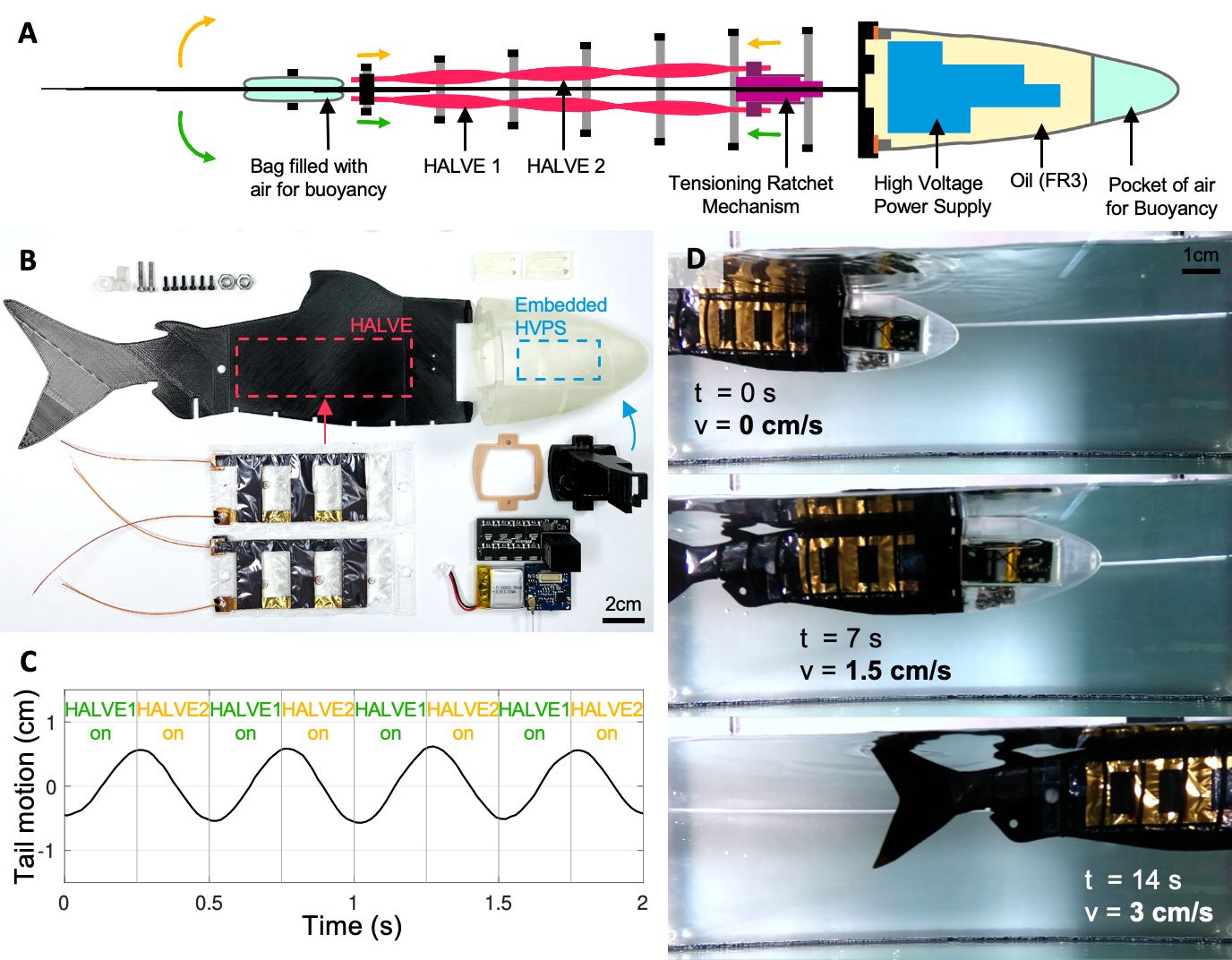}
    \caption{\textbf{Untethered artificial fish for underwater locomotion.} (\textbf{A})~The soft robotic fish assembly structure. The green and yellow arrows indicate the forces generated by the \hl{HALVE actuators}. (\textbf{B})~Components of the fish before assembly. Airbags for buoyancy and the ribs are not displayed. (\textbf{C})~Left and right oscillating motion measured at the tip of the tail. Indicated are also the antagonistic activation of the \hl{HALVE actuator}. (\textbf{D})~Swimming fish accelerating from a stationary position to reach a speed of \qty{3}{\centi\meter\per\second} after \qty{14}{\second}.}
    \label{fig:Fish}
\end{figure}

        We validated and tested the integrated actuator system in a fully untethered, self-contained system, namely the bio-inspired swimmer seen in Fig.~\ref{fig:Fish}. We prepared a $30 \times 35 \times 60$ \qty{}{\centi\meter} tank filled up to a height of \qty{30}{\centi\meter} with room temperature tap water. We placed the swimmer in the aquarium and actuated the \hl{HALVE actuators} antagonistically to observe forward thrust. Movie S6 shows the untethered artificial fish swimming in tap water with \qty{2}{\hertz} antagonistic actuation. We recorded the motion trajectory of the fishtail (Fig.~\ref{fig:Fish}C) and measured the performance of the fish by its forward swimming velocity. We observed a maximum swimmer velocity of \qty{3.8}{\centi\meter\per\second}, equivalent to 0.14 bodylength/second (BL/s) at \qty{2}{\hertz} antagonistic actuation. Aside from forward swimming velocity, we also characterized the efficiency of our swimmer by measuring the power consumption of the high voltage power supply during \qty{2}{\hertz} antagonistic actuation, which was on average \qty{0.6}{\watt}. On full charge, the battery of \qty{0.56}{\watt\hour} can therefore last about \qty{1}{\hour} during actuation. The swimmer also has an integrated internal measurement unit (IMU) and can be communicated with via \qty{433}{\mega\hertz} radio, which works underwater up to a depth of a few centimeters.

\section*{Discussion}
    We have shown a promising new actuator system called \hl{hydraulically amplified low voltage electrohydraulic (HALVE) actuator}. The key properties of the presented system are its low-voltage characteristic, high power density, easy implementation characteristic, and high robustness (self-clearing). \hl{The three}-layer design approach has enabled us to use high energy density dielectrics, drastically reducing the voltage requirements of Peano-HASEL actuators to 100s of volts. In some areas, such as peak strain rate and average specific power they are similar to the performance of mammalian skeletal muscle. \hl{HALVE actuators} possess several desirable features for implementation such as safe operation in proximity to humans, and the ability to operate in water. Due to the small size and high efficiency of the required power supplies, these actuators can easily be scaled to multichannel systems. This characteristic is in contrast to most electrostatic actuators, which are typically hard to implement. 
    
    One challenge with the current design is the durability of the heat seals, which typically only fuse the P(VDF-TrFE-CTFE) layers, not the outer structural shell. To address this problem, we have developed a manufacturing method that uses BoPVDF as the outer shell, which was used for the fish demonstrator. However, this new method comes with drawbacks, including a significant increase in material cost, required equipment, and manufacturing time and labor. Another challenge is the actuators' relatively low force density, which is typical for electrohydraulic actuators. Although \hl{HALVE actuators'} average specific power is in the range of natural muscle (\qty{50}{\watt\per\kilo\gram}), the maximum force density is currently lower. Force density could be improved by increasing the breakdown voltage of the P(VDF-TrFE-CTFE) thin films. The quality of our thin dielectric films is limited by the current manufacturing processes, which typically results in a dielectric strength of \qty{130}{\volt\per\micro\meter} (approx. half of manufacturer's specifications). However, moving manufacturing into clean rooms could improve the purity of the dielectrics.
    
    %Conclusion
    Moving forward, there are many opportunities to explore the application of more sophisticated high-energy density dielectrics to \hl{HALVE actuators} and to electrohydraulic actuators in general. Recently developed dielectrics such as BNT-NN/PVDF-HFP nanocomposites could further increase power density to outperform mammalian skeletal muscle~\cite{Xie2022UltraHighEnergyDensity}. Even smaller driving electronics, such as the ones shown by Ji et al.~\cite{ji2019autonomous} would enable biologically inspired autonomous robots with many independent channels. Other manufacturing methods for the deposition of the dielectric to create higher-quality thin films with high breakdown strength should be investigated. One promising technology is electrospraying, which has been successfully used to create very thin high-quality dielectric films for dielectric elastomer actuators~\cite{Weiss2016Electrospraying}. As it stands, the characteristics of \hl{HALVE actuators} shown in this study highlight their promise for next-generation untethered soft robotic systems.
    
    \FloatBarrier
    % \clearpage

\section*{Materials and methods}
    \subsection*{Actuator materials and geometry}
        %Characterization \hl{HALVE actuators}
        \hl{HALVE actuators} used for characterization were made of a single \qty{60}{\milli\meter} wide by \qty{17}{\milli\meter} tall pouch. This wide aspect ratio was chosen so that the structural film edge effects were as small as possible for better matching with the model~\cite{kellaris2019analytical}. The electrodes area on the pouch was chosen to be \qty{9}{\milli\meter} by \qty{59}{\milli\meter}, and did not cover the pouch width entirely, leaving a \qty{0.5}{\milli\meter} gap between the electrode and the pouch seal. This was done to avoid breakdown at the heat seal location, where the P(VDF-TrFE-CTFE) layer is thinner as a result of the heat sealing procedure. The top edge of pouch was instead sealed as close as possible to the electrode without overlapping it, as this is required for the formation of a zipping front. The sides of the pouch section not covered by the electrodes had triangular indents, to reduce the influence of edge effects~\cite{kellaris2019analytical}. A \qty{12.5}{\micro\meter} thick BoPET film was used as the structural shell, while the electrode was made of Aluminum (Fig. \ref{fig:HASEL_Performance}B).
        %Characterization HASEL actuators
        The HASEL actuators used for comparison purposes used the same pouch geometry and dimensions as the \hl{HALVE actuators} described above. A \qty{15}{\micro\meter} thick BoPET film was used as the solid dielectric and structural shell layer, while the electrodes were made from carbon ink.
        
        %Gripper actuators
        The \hl{HALVE actuators} used in the gripper demonstrator were made of three pouches, each measuring \qty{35}{\milli\meter} in width and \qty{15}{\milli\meter} in height, with electrode covering a \qty{34}{\milli\meter} by \qty{7.5}{\milli\meter} section of the pouch.
        A \qty{12}{\micro\meter} heat-sealable BoPET film was used as the structural shell (Hostaphan RHS12 provided by Mitsubishi Polyester Film GmbH). The heat-sealable characteristics produced a more reliable pouch seal compared to BoPET alone while maintaining similar mechanical properties.
        With this approach, the heat-sealable layer of the structural shell is melted, not allowing the use of electrode traces to connect the electrode to the outside of the pouch. For this reason, \qty{50}{\micro\meter} copper wires are used to connect the individual electrodes. The electrode was made of Chrome and Gold.
        %Fish actuators
        The \hl{HALVE actuators} used for the fish demonstrator were made of three \qty{31}{\milli\meter} by \qty{23}{\milli\meter} pouches, with an area of \qty{11.5}{\milli\meter} by \qty{31}{\milli\meter} covered by each electrode. 
        A \qty{12}{\micro\meter} BOPVDF film was used as the structural shell. P(VDF-TrFE-CTFE) bonds to this material when melted, producing a more reliable pouch seal compared to BoPET alone. The electrode was made of Chrome and Gold.
        All previously described \hl{HALVE actuators} had a \qty{5}{\micro\meter} dielectric layer was made of Poly (P) vinylidene fluoride (VDF), trifluoroethylene (TrFE), 1,1-chlorotrifluoroethylene (CTFE) powder from Piezotech-Arkema (Piezotech RT-TS, P(VDF-TrFE-CTFE) terpolymer). All HASEL and \hl{HALVE} devices used Envirotemp FR3 from Cargill as the dielectric liquid.
        
    \subsection*{Actuator manufacturing}
    % Strucural shell and electrode   
        BoPET structural shell and aluminum electrode were obtained by processing a survival blanket (Forclaz Decathlon). The Chrome-Gold electrode was produced by vapor deposition. First, a \qty{5}{\nano\meter} layer of chrome was deposited to improve adhesion, followed by a \qty{60}{\nano\meter} layer of gold. The dielectric layer of the \hl{HALVE actuators} was produced by blade casting a \qty{150}{\micro\meter} layer of 14\% wt. P(VDF-TrFE-CTFE) in Methyl Ethyl Ketone (MEK) solution onto the structural shell and electrodes. For this, a TQC Sheen Automatic Film Applicator (AB3652) with a Film Applicator Block from Zehntner (ZUA 2000) was used. After casting, P(VDF-TrFE-CTFE) thin films were annealed at \qty{102}{\celsius} for \qty{2}{\hour}. Further information on the casting can be found in the supplementary materials.
        %sealing + filling
        The \hl{HALVE actuators'} oil pouches were created by heat-sealing. A standard 3D printer (Prusa MK3S) was used to print a single line of ABS filament at \qty{295}{\celsius} onto a \qty{25}{\micro\meter} Kapton sheet which transferred the heat to the actuator films below. The amount of dielectric oil was chosen to be 95\% of the theoretical maximum cylindrical volume of the fully deformed pouch. The Peano-HASEL actuators were produced following methods described in the literature \cite{mitchell2019toolkit}, with the only difference being that they were sealed with the technique described above, but by printing PLA at \qty{220}{\celsius} to activate the heat-seals. More details on the manufacturing can be found in the supporting information.

    \subsection*{Fish manufacturing and measurements}
        The fish measures \qty{28}{\centi\meter} in length and is composed of two sections: First, the head (Fig. \ref{fig:Fish}A) is made of a rigid and semi-transparent watertight shell that was 3D printed with selective laser sintering (SLS). The head contains the electronics and was filled with dielectric oil (Envirotemp FR3) to protect the electronics in the event of water entering a cavity. Second, the body is made of a flexible 3D-printed skeleton of polypropylene (PP) onto which the \hl{HALVE} actuators were attached. A tensioning ratchet system was used to tighten the actuators individually to conform to any deviations in manufacturing. The structure contains air pouches that balance the fin and the head and allows the system to be a surface swimmer. The flexible body of the fish demonstrator onto which the actuators were attached was printed from black PLA with a PRUSA~MK3S. The head of the fish was printed from Formlabs Durable Resin with a Formlabs~3. We measured the speed and motion of the fishtail with a camera recording the fish profile from the top with a reference grid below the water tank. The energy consumption of the fish electronics unit was measured by supplying power to the fish with a DC bench power supply and recording the average current.
    
    %Measurements    
    \subsection*{Force-strain characterization}
        The actuators were characterized with a setup in which the actuators' displacement is measured, while forces of various magnitudes are applied to the actuators via a set of weights. The displacement was measured using a Baumer OM70-11216521 distance sensor. Low voltage control inputs were generated via a custom LabView VI and a DAQ (NI USB-6343) from National Instruments, and subsequently amplified with a Trek 20/20C-HS high voltage amplifier. In order to determine the quasi-static force-strain relation of the actuators the actuation strain was measured while applying forces between \qty{0.2}{\newton} and \qty{5}{\newton}. The voltage signal used for the force-strain characterization was a bipolar square signal at a frequency of \qty{0.1}{\hertz}. In order to obtain statistically relevant data, the strain of four subsequent actuation cycles was recorded, and the mean was calculated. In order to avoid transient phenomena at actuation the data of the first second after application of the voltage was not considered. This process was repeated at different actuation voltages.
 
    \subsection*{Strain rate characterization}
        The voltage signal used for the force-strain characterization was a bipolar square signal at a frequency of \qty{0.1}{\hertz}. A \qty{0.2}{\newton} force was applied to the actuators, and the produced strain was measured with the previously described setup. The actuation time was quantified as the time required to reach the mean steady-state strain value after voltage application. The strain rate was obtained by dividing the mean steady-state strain by actuation time. In order to obtain statistically relevant data, each measurement was acquired four times. This process was repeated at different actuation voltages.
    
    \subsection*{Durability test}
        Durability tests were performed on single pouch actuators with the geometry shown in Fig. \ref{fig:hasel_model}Aii. Fig.~\ref{Sfig:measurement}B shows a durability test over 2500 Cycles at \qty{1}{\hertz} lifting a \qty{200}{\gram} weight actuated by a bipolar \qty{800}{\volt} signal supplied by a high voltage lab power supply (TREK 20/20C-HS). The strain was recorded with the setup shown in Fig.~\ref{Sfig:measurement}A. During durability tests, the actuators' strain degraded a few percent over time with repeated actuation which is shown in Fig.~\ref{Sfig:measurement}B. This performance degradation varied across individual actuators, suggesting that manufacturing variations like contaminants in the P(VDF-TrFE-CTFE) layer or dielectric oil, humidity levels, or degradation of the sealing lines / slight leaking of oil play a role in this phenomenon.

\section*{Acknowledgments}
    We thank Piezotech-Arkema for providing the P(VDF-TrFE-CTFE) polymer powder and P. Werle for providing samples of dielectric oils. We thank M. Riner-Kuhn for his help in designing and manufacturing the high-voltage power supplies. We thank B. Böck, V. Kannan, M. Müller, and M. Trassin for advice on dielectric material theory and measurements. Lastly, we thank ScopeM for their support and assistance in this work. \textbf{Funding:}~This work was mostly funded by a generous donation from Credit Suisse to the ETH Foundation to create a new chair for Robotics at ETH Zurich. S.G. and T.B. were partially funded by the Zurich Heart project \#2022-FS-320 titled "Electrostatically actuated biomimetic artificial muscles for support or replacement of physiological functions such as pumping of the heart". A.K. was funded by the NCCR Robotics Grant titled "Versatile, affordable, and easy-to-use robotic grippers". A.K. is partially funded by a Swiss Government Excellence Scholarship. M.M. and A.K. are partially funded by the ETH Grant \#ETH-17 22-1 titled "Optimizing Fluidic Soft Robots with Differentiable, Multiphysics-Informed Neural Networks". M.M. is partially funded by a doctoral fellowship from the ETH AI Center.  M.M. is partially funded by the RobotX Research program \#RX-03-22 titled "LPDR - Learning Physics-based Dense Representations for Robotic Manipulation of Soft and Articulated Objects". \textbf{Author contributions:}~S.G., E.V., R.H., and R.K. designed the low-voltage electrohydraulic actuator and developed fabrication methods. E.V. and S.G. fabricated actuators, designed experiments, and collected data. A.K. and S.G. designed and tested the miniature high-voltage power supplies. E.V. designed and manufactured the untethered gripper. S.G. designed and manufactured the fish demonstrator and performed the dielectric material measurements. \hl{T.B. advised S.G. and E.V. on the development and manufacturing of actuators and robotic systems.} M.M. wrote the model optimization algorithm for the actuator energy density calculations. R.K. conceptualized and supervised the research. All authors contributed to the manuscript and approved the final draft. \textbf{Competing interests}~None. \textbf{Data and materials availability:} All data needed to support the conclusions of this manuscript are included in the main text or supplementary materials. Contact R.K. for further materials.

\section*{Supplementary Material}
    \subsection*{Summary}
    Materials and Methods

    Text S1. Mean value theorem for effective dielectric constant.

    Fig. S1. Manufacturing of structural shell and electrode from Survival Blanket.

    Fig. S2. Manufacturing of BoPVDF structural shell and gold electrode with additive manufacturing.

    Fig. S3. Blade casting of a thin film of P(VDF-TrFE-CTFE) onto BoPET/Aluminum structural shell and electrode.

    Fig. S4. Sealing and filling of the actuator.

    Fig. S5. Focused ion beam scanning electron microscope (FIB-SEM) and Atomic force microscope (AFM) measurements.

    Fig. S6. Characterization setup, \hl{HALVE} durability test, and actuation characteristics.

    Fig. S7. Average and peak specific power of \hl{HALVE} devices

    Fig. S8. Strain data from a \hl{HALVE} actuated at 1300V while lifting 200g.

    Fig. S9. Experimental actuation and regression results.

    Fig. S10. \hl{HALVE}s and power supply system of the untethered robotic fish.

    Fig. S11. Design of the compact high-voltage power supply (HVPS).

    Fig. S12. Schematic of power supply.

    Fig. S13. Footprint and cost of high voltage MOSFETs.

    Tab. S1. Comparison between high voltage switching components.

    Tab. S2. Power supply components list.

    Movie S1. Demonstration of \hl{HALVE device} actuation

    Movie S2. Touching a \hl{HALVE actuator} during activation

    Movie S3. Three-pouch \hl{HALVE} device operating in air and submerged in tap water

    Movie S4. Untethered \hl{HALVE} actuator system - Lifting its power supply

    Movie S5. Untethered gripper grasping a PLA block

    Movie S6. Untethered artificial fish driven by two antagonistic \hl{HALVE actuators} swimming in tap water

    References~\cite{Rumley2022ChargeRetention,cranmer2020discovering,MOSFET_datasheet,Opto_datasheet}
% \clearpage
\bibliographystyle{Science}
\bibliography{scibib}

\FloatBarrier

    \FloatBarrier

\FloatBarrier

\FloatBarrier 

\FloatBarrier

\FloatBarrier\clearpage

% %%%%%%%%%%%%%%%%%%%%%%%%%%%%%%%%%%%%%%%%
\setcounter{figure}{0} % restart figure counter for supplementary figures
\makeatletter 
\renewcommand{\thefigure}{S\@arabic\c@figure}
\renewcommand{\thetable}{S\@arabic\c@table}
\makeatother
%%%%%%%%%%%%%%%%%%%%%%%%%%%%%%%%%%%%%%%%%

\newpage
\section*{Supplementary materials}
\subsection*{Materials and Methods}
\subsubsection*{Structural shell and electrode fabrication}
\begin{figure}[h]
        \centering
        \includegraphics[width = 0.9\textwidth]{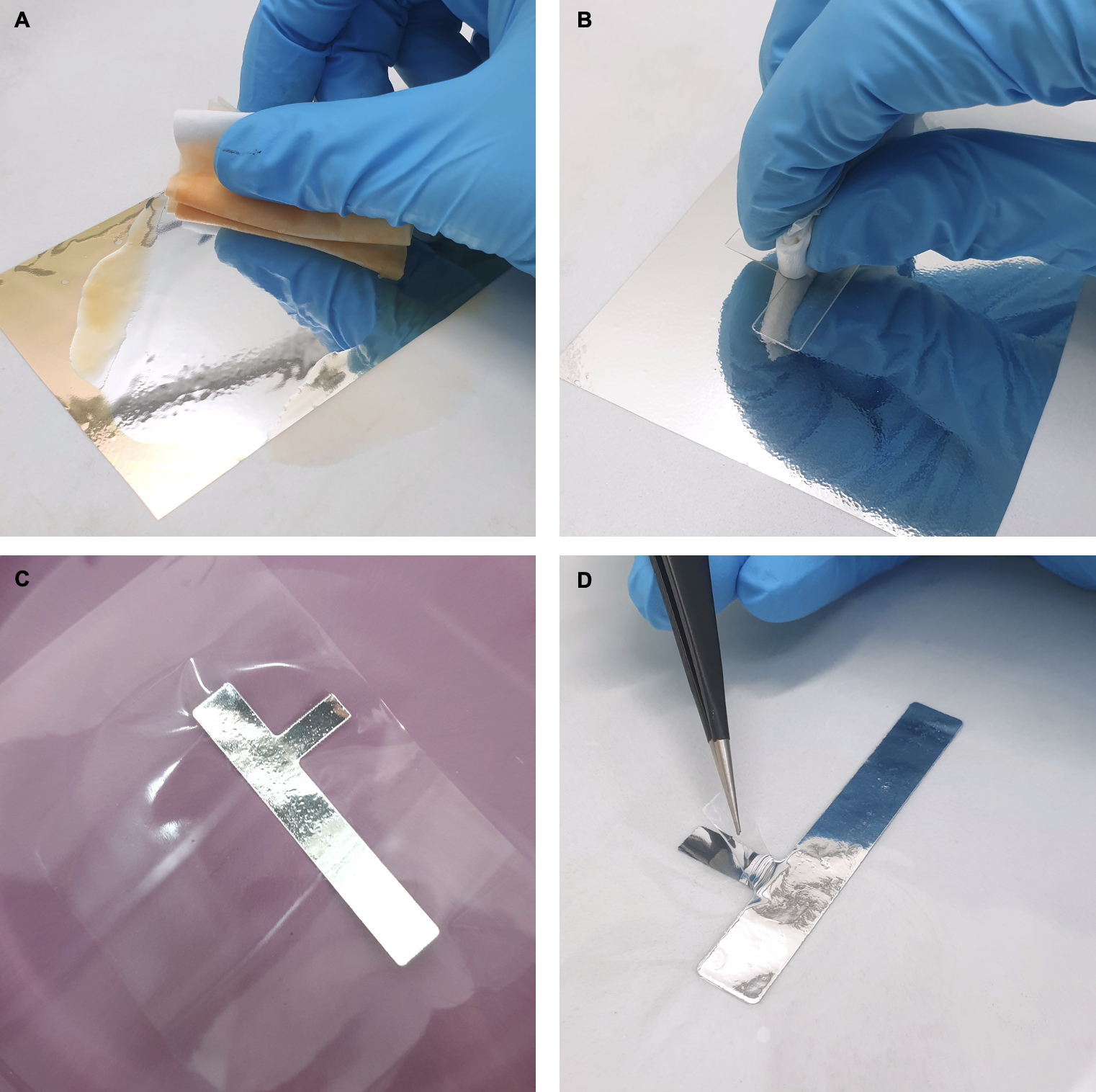}
        \caption{\textbf{Manufacturing of structural shell and electrode from Survival Blanket.} (\textbf{A})~Removing of the Polyethylene (PE) layer covering the Aluminum with acetone. (\textbf{B})~After the Aluminium is exposed, a previously precisely cut masking tape is pressed onto the film. (\textbf{C})~The film is submerged in a solution of potassium hydroxide to dissolve the Aluminum which is not covered by the mask. (\textbf{D})~Removing the masking tape with some acetone.}
        \label{figS:electrodeManufacturing}
    \end{figure}

    BoPET structural shell and aluminum electrode were obtained by processing a Mylar emergency blanket (Decathlon FORCLAZ Einweg-Rettungsdecke). These blankets are composed of a thin aluminum layer sandwiched between a PET and a PE layer. 
    The PE layer of the blanket was dissolved with acetone, exposing the aluminum (Fig. \ref{figS:electrodeManufacturing}A). The Aluminum layer then was masked with laser-cut Scotch tape (PP) (Fig. \ref{figS:electrodeManufacturing}B) and etched with a 2\% wt. potassium hydroxide solution (Fig. \ref{figS:electrodeManufacturing}C). After etching, the shells were rinsed under running tap water to remove the remaining potassium hydroxide. Acetone was then used to dissolve the glue of the Scotch tape after etching (Fig. \ref{figS:electrodeManufacturing}D). Once the masks were removed, the aluminum electrodes were wiped with acetone to remove the remaining adhesive, finishing the structural shell and electrode fabrication.

    The Chrome-Gold electrode was produced by e-beam evaporation (VERA 450 H VTD). In order to vapor deposit material only at the electrode location, laser-cut acrylic masks were used to mask the structural shells. A \qty{5}{\nano\meter} layer of Chrome was then first deposited on the structural shell at a rate of \qty{0.2}{A\per\second} to improve adhesion, followed by a \qty{60}{\nano\meter} layer of Gold deposited at a rate of \qty{1}{A\per\second}.
    For the gripper demonstrator, the electrodes of the actuator were connected to the outside of the pouch with \qty{50}{\micro\meter} thick copper wire, which was connected to the electrode with conductive copper tape prior to P(VDF-TrFE-CTFE) coating. 
    
\subsubsection*{P(VDF-TrFE-CTFE) coating}
\begin{figure}[h]
        \centering
        \includegraphics[width = 0.9\textwidth]{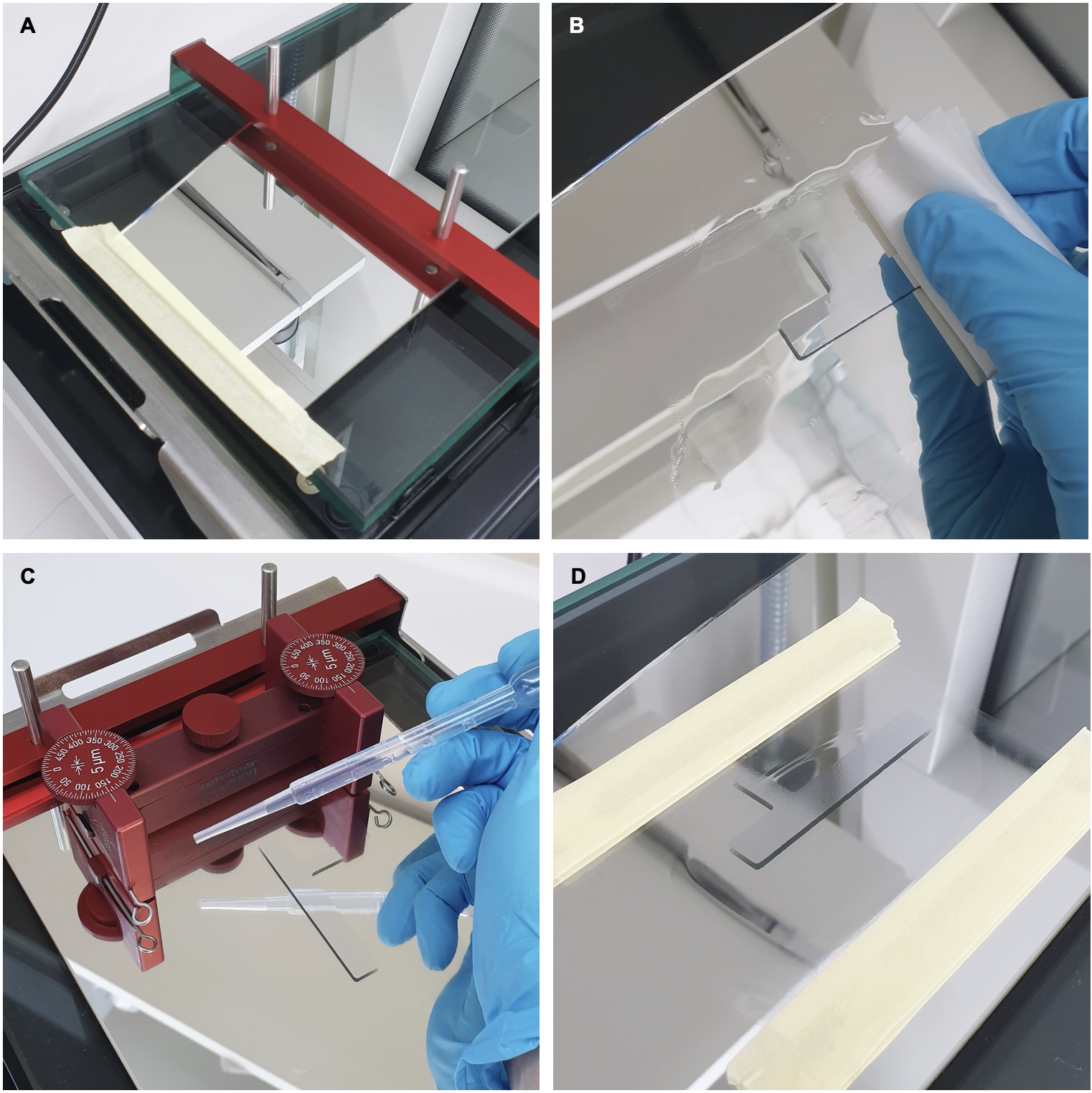}
        \caption{\textbf{Blade casting of a thin film of P(VDF-TrFE-CTFE) onto BoPET/Aluminum structural shell and electrode.} (\textbf{A})~Mirror is taped to a blade coating machine. (\textbf{B})~Evaporating acetone is used to place the structural shell flat to the mirror. (\textbf{C})~Blade casting block is placed onto the applicator and P(VDF-TrFE-CTFE) solution is applied. (\textbf{D})~After application the structural shell is taped to the mirror to prevent warping during the annealing process.}
        \label{figS:DielectricManufacturing}
    \end{figure}

\begin{figure}[h]
\centering
\includegraphics[width = \textwidth]{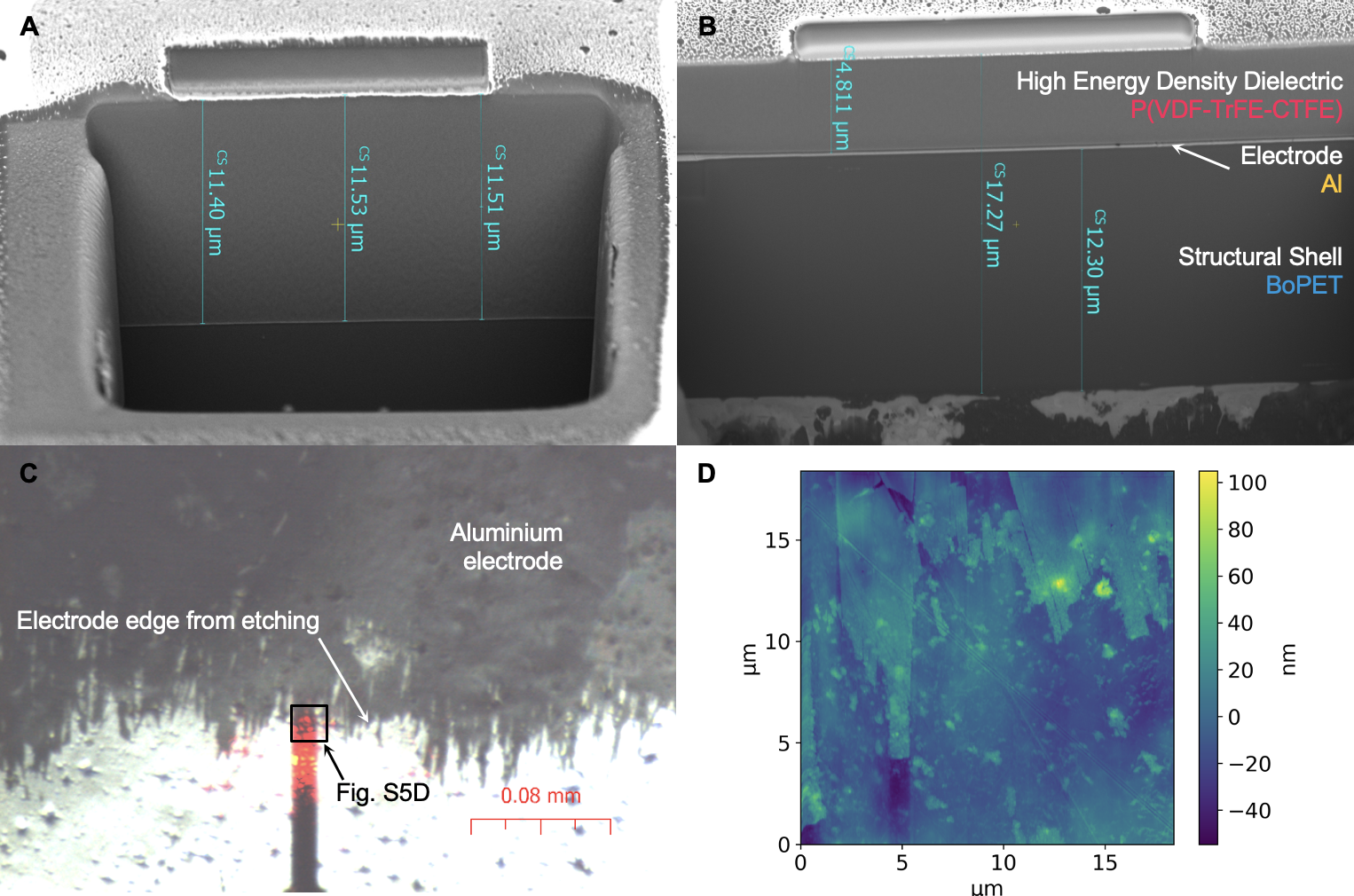}
\caption{\textbf{Focused ion beam scanning electron microscope (FIB-SEM) and Atomic force microscope (AFM) measurements.}
(\textbf{A})~FIB-SEM cross-sectional view of a BOPET structural shell and Aluminium electrode which was blade-coated with a \qty{300}{\micro\meter} layer of P(VDF-TrFE-CTFE) and MEK solution. After annealing the P(VDF-TrFE-CTFE) layer is shown to be approximately \qty{11.5}{\micro\meter} thick.
(\textbf{B})~FIB-SEM cross-sectional view of a BOPET structural shell and Aluminium electrode which was blade-coated with a \qty{150}{\micro\meter} layer of P(VDF-TrFE-CTFE) and MEK solution. The P(VDF-TrFE-CTFE) layer after annealing measures approximately \qty{5}{\micro\meter} in thickness.
(\textbf{C})~Optical microscope view of parts of an Aluminium electrode edge produced by etching.
(\textbf{D})~Atomic force microscope (AFM) measurements of the Aluminium electrode surface. The surface is very consistent, with nanometric scale (\qty{1}{\nano\meter}-\qty{100}{\nano\meter}) roughness.}
\label{figS:materialThickness}
\end{figure}

    The solid dielectric used for \hl{HALVE actuators} was P(VDF-TrFE-CTFE) from Piezotech-Arkema (Piezotech RT-TS, P(VDF-TrFE-CTFE) terpolymer). In preparation for the P(VDF-TrFE-CTFE) coating, the actuators' shells and a glass substrate (a rectangular mirror) were wiped with acetone to remove dust particles and other contaminants. The actuator's structural shell and connected electrode were then adhered electrode-side up to the glass substrate by laying them on an acetone pool, which was then gently squeezed out from below the shells, pulling the films flat on the glass plate (Fig.~\ref{figS:DielectricManufacturing}B). A P(VDF-TrFE-CTFE) layer was then applied on top of the shells by blade casting a solution of P(VDF-TrFE-CTFE) and Methyl Ethyl Ketone (2-Butanone) in a \qty{14}{\percent}  to \qty{86}{\percent} ratio by weight (Fig.~\ref{figS:DielectricManufacturing}C). The equipment used for casting was a TQC Sheen Automatic Film Applicator (AB3652) and a Film Applicator Block from Zehntner (ZUA 2000). In order to obtain a P(VDF-TrFE-CTFE) layer roughly \qty{5}{\micro\metre} thick, the film applicator block was set to lay a \qty{150}{\micro\metre} film of P(VDF-TrFE-CTFE) and MEK solution. The coated shells were then left to dry in air for \qty{2}{} minutes, during this time they were taped to the substrate using masking tape to prevent subsequent warping in the oven (Fig.~\ref{figS:DielectricManufacturing}D). The coated sheet was then annealed in an oven at \qty{102}{\celsius} for \qty{2}{} hours. The ratio between cast layer thickness and final dried layer thickness was measured to be approx. 1:30 for a casting thickness of \qty{150}{\micro\meter} and approx. 1:25 for a casting thickness of \qty{300}{\micro\meter}. A Focused Ion Beam Scanning Electron Microscope (FIB-SEM) (ThermoScientific Helios 5 UX) was used to take a picture of the cross-section of the PET-Aluminum-P(VDF-TrFE-CTFE) sandwich. Three samples (different actuators) were measured. Two were coated with \qty{150}{\micro\meter} layers of P(VDF-TrFE-CTFE) and one with \qty{300}{\micro\meter}, which shrunk to \qty{5}{\micro\meter} and \qty{11.5}{\micro\meter} respectively after evaporation and annealing (Fig. \ref{figS:materialThickness}A and Fig. \ref{figS:materialThickness}B). The smaller thickness reduction ratio suggests a more porous P(VDF-TrFE-CTFE) layer for thicker P(VDF-TrFE-CTFE) coatings, which could affect durability and breakdown strength negatively. Reducing the porosity might be achieved by increasing the air evaporation time before putting the samples into the oven to slow down evaporation.
    
    \subsubsection*{Pouch sealing}
    In order to produce the oil pouches, \hl{HALVE actuators} were heat-sealed. The two actuators' halves were first stacked by aligning the electrodes. Then the actuator was positioned on a 3D printer's bed (Prusa MK3S) between two \qty{25}{\micro\meter} Kapton films. The actuator's pouches were then produced by printing a single layer of hot filament at the sealing-line location (see Fig.~\ref{figS:FillingAndSealing}A).
    In order to seal actuators with BoPET and Hostaphan structural shells, ABS filament was printed at a nozzle temperature of \qty{293}{\degreeCelsius}. In the case of the BOPVDF structural shell PETG filament was used, at a nozzle temperature of \qty{240}{\degreeCelsius}. HASEL actuators were sealed using PLA filament, at a nozzle temperature of \qty{220}{\degreeCelsius}.
    For all of the above the printer bed's temperature was set to \qty{30}{\degreeCelsius}.
    
    \subsubsection*{Oil filling}
\begin{figure}[h]
        \centering
        \includegraphics[width = 0.9\textwidth]{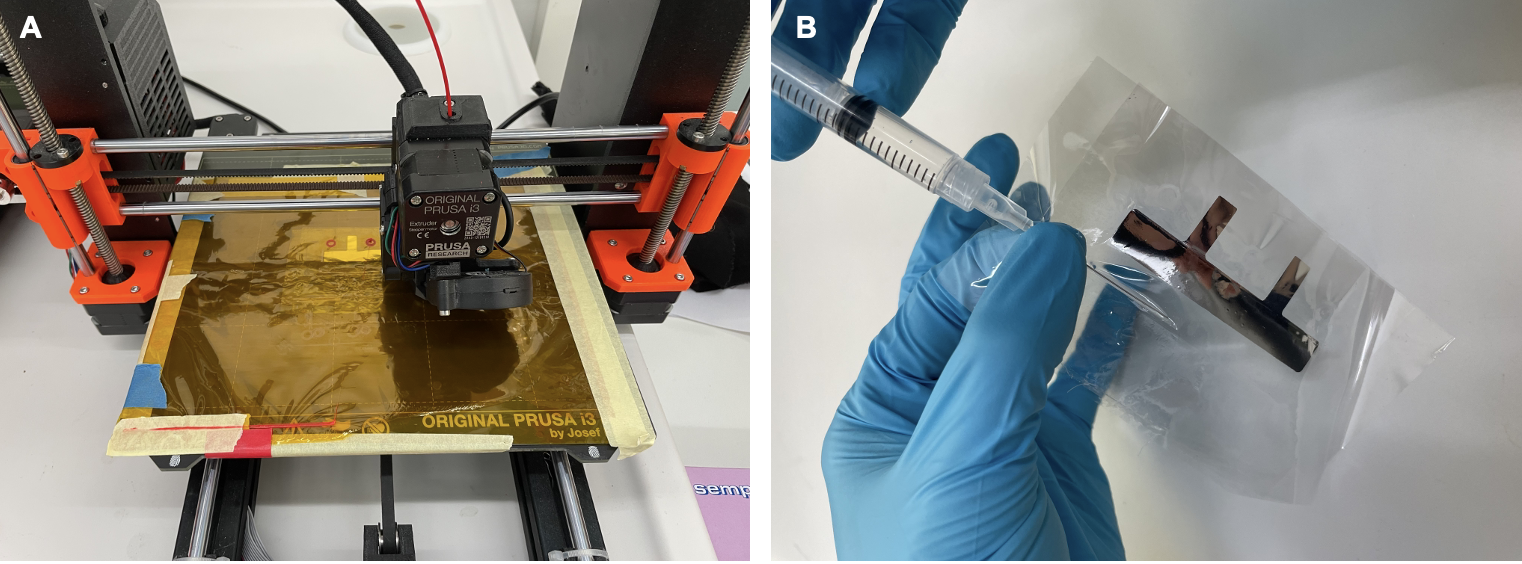}
        \caption{\textbf{Sealing and filling of the actuator.} (\textbf{A})~Two coated thin films are placed between two Kapton sheets for sealing. One layer of filament is printed onto the Kapton sheet to transfer heat to the actuator thin films. (\textbf{B})~Actuator filling with a syringe and blunt tip through a filling port.}
        \label{figS:FillingAndSealing}
    \end{figure}

    Once the pouch had been sealed, it was filled with dielectric oil. The amount of oil used was \qty{95}{\percent} of the theoretical cylindrical volume formed when the actuator is fully actuated. We tested different liquid dielectrics (Midel 7131, Shell GTL S4, Envirotemp FR3) and chose Envirotemp FR3 from Cargill. It was chosen as it has been used successfully in previous hydraulically amplified electrostatic actuators and demonstrated the best chemical compatibility with P(VDF-TrFE-CTFE). The pouch was filled through the filling port using a syringe with a long and thin, blunt point needle Fig. (\ref{figS:FillingAndSealing} B). Once the pouch was filled, the actuator was positioned between two Kapton sheets. The filling port was then heat sealed by applying light pressure with a soldering iron set at \qty{275}{\degreeCelsius} for BoPET and Hostaphan structural shells, \qty{240}{\degreeCelsius} for BOPVDF structural shells, and \qty{225}{\degreeCelsius} for HASEL actuators. The excess actuator's shell was then trimmed, and the mounting holes were created with a revolver hole punch.

    \subsubsection*{Strain rate and specific power derivation}
\begin{figure}[h]
        \centering
        \includegraphics[width = \textwidth]{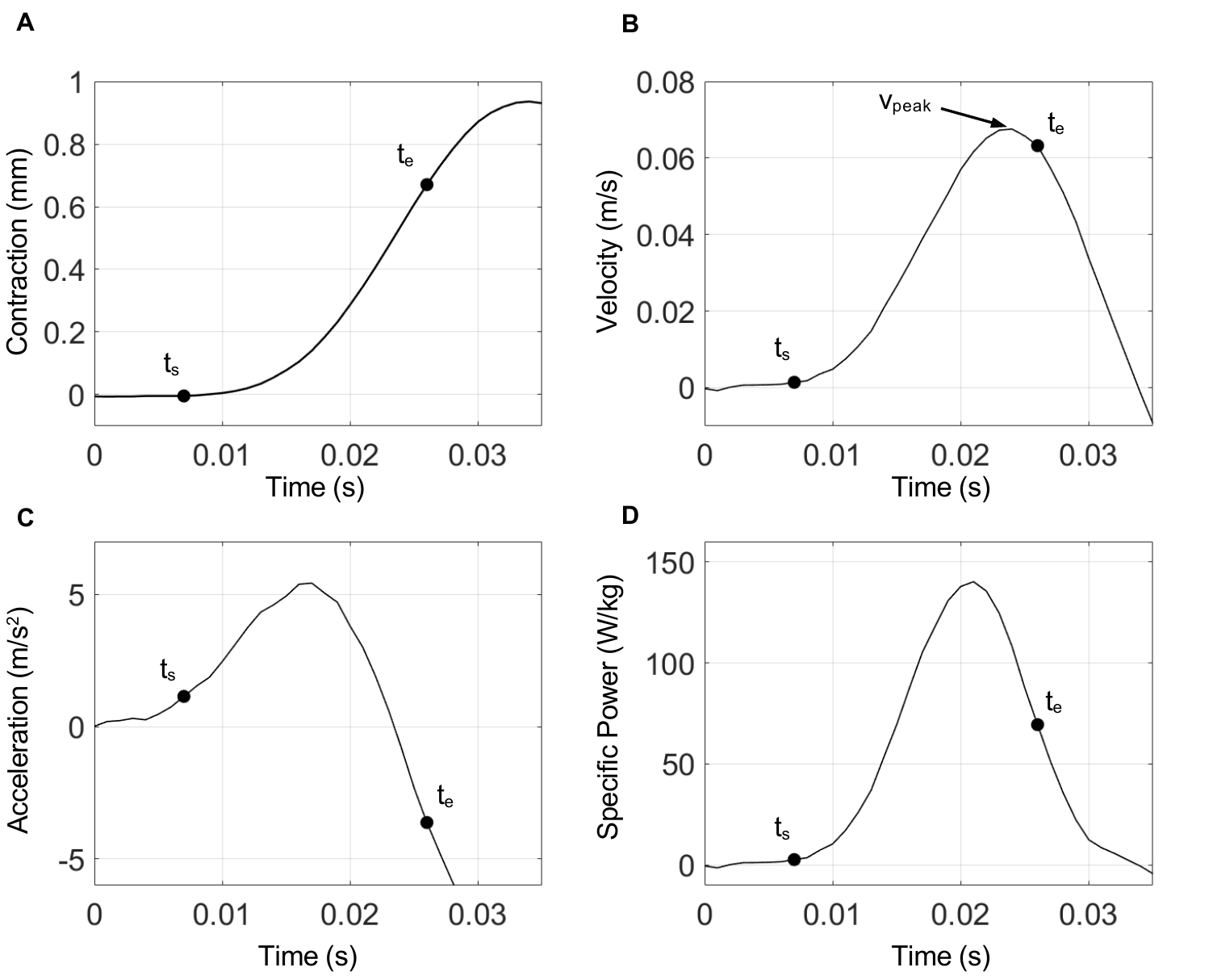}
        \caption{\textbf{Strain data from a \hl{HALVE actuator} driven at $\mathbf{1300V}$ while lifting $\mathbf{200g}$.} (\textbf{A})~We use the same measurement method as Kellaris et al.~\cite{kellaris2018peano}. Measured contraction after a voltage step input is applied to the actuator. A Savitsky-Golay filter was applied to the data. (\textbf{B})~Contraction velocity curve. The maximum value of this curve was used to calculate the actuation peak strain value with Eq.~\ref{eq:PeakStrainRate}. (\textbf{C})~Contraction acceleration curve. Both contraction velocity and acceleration were obtained by taking the derivative of the Savitsky-Golay polynomials. (\textbf{D})~Specific power curve, which was calculated with Eq.~\ref{eq:P_sp}.}
        \label{figS:specificPowerExplanation}
    \end{figure}

    Peak strain rate and peak and average specific power of \hl{HALVE actuators} and Peano-HASEL actuators were determined following the methodology described by Kellaris et al.~\cite{kellaris2018peano}. The characterization setup shown in \ref{Sfig:measurement} was used to record the actuators' contraction in response to a \qty{0.1}{\hertz} square wave. The voltage polarity was reversed at each actuation. To smooth the measured contraction data a Savitsky-Golay filter was applied to it (Fig.~\ref{figS:specificPowerExplanation}A). The filter used a Third-order polynomial fit and a frame length of \qty{17}{}. Contraction speed (Fig.~\ref{figS:specificPowerExplanation}B) and acceleration ((Fig.~\ref{figS:specificPowerExplanation}C)) were then calculated by taking the derivatives of the Savitsky-Golay polynomials. For all the calculations described below, only the period from initial movement ($t_{s}$) to reaching equilibrium contraction ($t_{e}$) was considered.
    
    The peak strain rate was calculated as follows:
    \begin{equation} \label{eq:PeakStrainRate}
        Peak Strain Rate = \frac{v_{peak}}{L_{actuator}} \cdot 100\%
    \end{equation}
    Where $v_{peak}$ is the peak contraction velocity, and $L_{actuator}$ is the actuator length.

    To calculate specific power, first, the force acting on the weight hanging from the actuator was determined:
    \begin{equation} \label{eq:F_net_1}
        F_{net}(t) = m_{weight} \cdot a(t) = F_{act}(t) - F_{g}(t)
    \end{equation}
    \begin{equation} \label{eq:F_net_2}
        \begin{split}
        F_{act}(t) & = F_{net}(t) + F_{g}(t) \\
                   & = m_{weight} \cdot a(t) + m_{weight} \cdot a_{g} \\
                   & = m_{weight} \cdot (a(t) + a_{g} )
        \end{split}
    \end{equation}
    Where $a_{g}$ is the acceleration due to gravity and $m_{weight}$ is the mass of the hanging weight.
    Specific power (Fig.~\ref{figS:specificPowerExplanation}D) was then derived by dividing the product of force and actuation speed by actuator mass:
    \begin{equation}  \label{eq:P_sp}
        P_{sp}(t) = \frac{P(t)}{m_{actuator}} = \frac{F_{act} \cdot v(t)}{m_{act}}
    \end{equation}
    Average specific power was instead derived by integrating the specific power between $t_{s}$ and $t_{e}$ to obtain total specific work, and finally dividing by the change in time.
    \begin{equation}  \label{eq:P_sp_avg}
        P_{sp-avg}(t) = \frac{W_{sp-total}}{t_{e}-t_{s}}
    \end{equation}

\subsubsection*{Box-constrained optimization and system identification}
\begin{figure}[h]
        \centering
        \includegraphics[width = \textwidth]{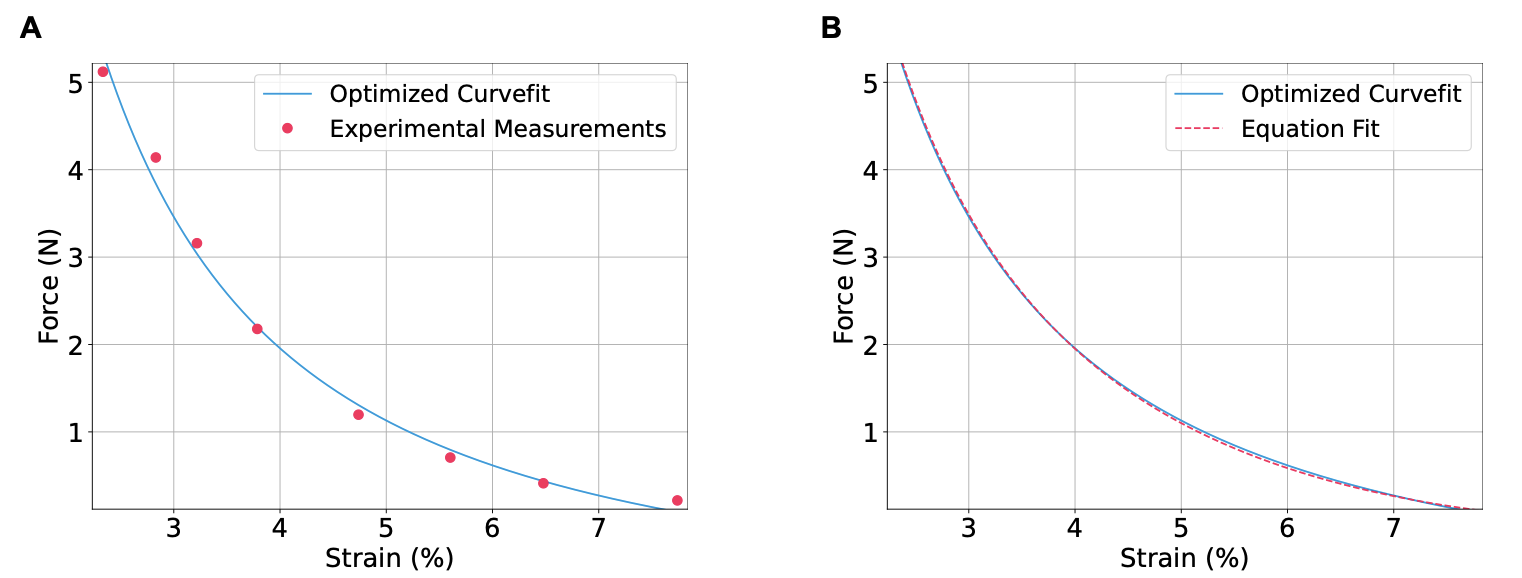}
        \caption{\textbf{Experimental actuation and regression results.} (\textbf{A})~Fitting a parametric curve to experimental HASEL actuation data of $1300V$. (\textbf{B})~Symbolic regression result and equation that fits the experimentally fitted curve.}
        \label{figS:curvefit}
    \end{figure}

    To integrate the area under the force/strain measurements, we used a box-constrained optimization algorithm to estimate a fitting curve. The eight relevant parameters from the force/strain equation derived by Kellaris et al.~\cite{kellaris2019analytical} that were optimized were width $w$ of the pouch, thickness $t$ of the pouch, $\varepsilon_0$, $\varepsilon_r$, applied voltage $V$, $\alpha_0$, $L_e$, $L_p$. We assumed physically realistic minimum and maximum values for each parameter and searched for a combination that best fits our experimental data. We additionally optimized the half-central angles $\alpha$ that parametrize the relation between force and strain. The parameters were normalized and clamped between $[0,1]$ during optimization based on the minimum and maximum values that we set. An optimized curvefit, for \textit{e.g.} \qty{1300}{\volt}, can be seen in Fig.~\ref{figS:curvefit}.
    % The upper/lower bounds used are:
    % w_ = [1e-2, 1e-1]
    % t_ = [1e-6, 1e-5]
    % eps_0_ = [8e-12, 1e-11]
    % eps_r_ = [1e1, 1e2]
    % V_ = [1e2, 2e3]
    % alpha_0_ = [1e-2, 1e0]
    % L_e_ = [1e-3, 1e-1]
    % L_p_ = [1e-3, 1e-1]
    Since we were interested in an analytical expression of force as a function of strain, we performed symbolic regression using PySR \cite{cranmer2020discovering} on the optimized curve-fit, since this curve-fit provides us with more sampled data points for the regression than the original experimental data. Running the symbolic regression gave us several analytical expressions with increasing complexity, from which we chose Eq.~\ref{eq:eqfit}. We observed that the range of data points given to the symbolic regression played a crucial role. The stress rapidly decayed to zero with larger strains, and if we were to sample data points uniformly on a strain domain between, for example, $[0, 100]\%$, then the majority of stresses would be zero, whereas the important datapoints at low-strain would become outliers. This would be harder to symbolically regress, and additionally, the high-strain regime was not of interest to us. Our shown results were achieved by limiting the strain domain to approximately $[1, 10]\%$. This does, however, imply that the predictive capabilities of the analytical expression failed at larger strains (and likely also at very small strain values).
    %'((inv(sin((x0 + -0.22899766) * 0.16584007)) * 2.8149376) + -2.8581402)'
    \begin{equation}
        F (\varepsilon) = \frac{2.81}{\sin \left(0.17 (\varepsilon - 0.23) \right)} - 2.86
        \label{eq:eqfit}
    \end{equation}

\subsubsection*{Miniature high-voltage power supply}
\begin{figure}[h]
\centering
\includegraphics[width = \textwidth]{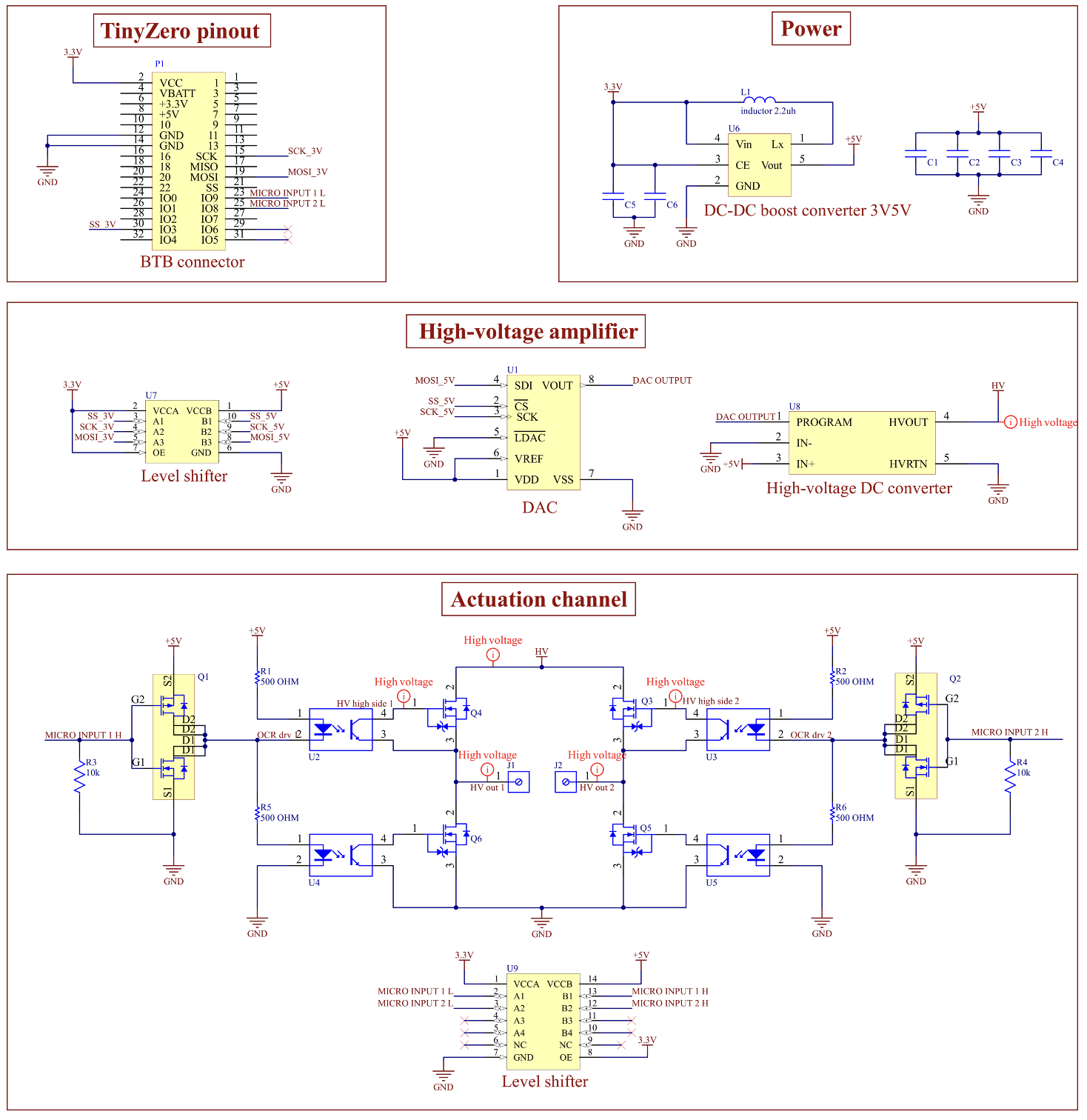}
\caption{\textbf{Schematic of power supply.} Detailed electrical schematic of the high-voltage power supply.}
\label{figS:schematic}
\end{figure}

    The electrical schematic for the modular power supply, designed with ALTIUM DESIGNER software (version 22.7.1), is shown in Fig.~\ref{figS:schematic}. A list of the components shown in the schematic is provided in Table~\ref{tabS:listOfComponents}. The TinyZero from TinyCircuits was selected as the processor board. The TinyZero is equipped with a 32-bit Atmel SAMD21 ARM Cortex M0+ processor, which is integrated into a compact \qty{20}{\milli\meter} x \qty{20}{\milli\meter} board weighing only \qty{1.4}{\gram}. Additionally, the board includes a Bosch BMA250 3-axis accelerometer. Its small size and light weight make it a good choice for autonomous robotic systems, and it can be easily expanded with other stackable TinyShield boards. For example, a TinyZero board may be combined with a long-range radio module or a short-range Bluetooth module to enable wireless communication capabilities. The high-voltage power supplies are powered by a lithium-ion polymer battery (\qty{150}{\milli\ampere\hour}) with dimensions of \qty{20}{\milli\meter} x \qty{20}{\milli\meter} x \qty{5}{\milli\meter} and weight of \qty{3.77}{\gram}. A step-up DC-DC converter is used to supply regulated power to the high-voltage DC amplifier. We chose ultra-miniature unipolar, regulated high voltage DC to DC converters by \textit{HVM Technology} as the high voltage amplifier. Fig.~\ref{figS:electronics}A shows the top views of two high-voltage power supplies with different power capacities. We used the \textit{nHV Series} (nHV0510), which measures \qty{11.4}{\milli\meter} x \qty{8.9}{\milli\meter} x \qty{9.4}{\milli\meter}. For the low-power HV amplifer (\qty{0.1}{\watt}), we used the \textit{nHV Series} (nHV0510), measuring \qty{11.4}{\milli\meter} x \qty{8.9}{\milli\meter} x \qty{9.4}{\milli\meter}. For the more powerful HV amplifier employed in the demonstrators, we utilized a \qty{0.5}{\watt} DC-DC converter from the \textit{UMHV Series} (UMHV0510) with dimensions of \qty{12.7}{\milli\meter} x \qty{12.7}{\milli\meter} x \qty{12.7}{\milli\meter}. Both power supplies can deliver up to \qty{1}{\kilo\volt} output signal and feature a high-impedance programming input. An external 12-bit Digital-to-Analog Converter (DAC) was incorporated into the circuit to achieve precise control over the input voltage of the high-voltage amplifier. A level shifter is used to translate the signals from 3.3V to 5V, enabling the connection between the DAC and the microcontroller through the Serial Peripheral Interface (SPI). Each actuation channel comprises four MOSFETs in a full H-bridge configuration: One MOSFET pair charges the actuator while the other pair discharges it. This configuration allows for bipolar actuation of each \hl{HALVE actuator}. The MOSFETs feature a maximum drain-source voltage $V_{DS}$ of 950V and are driven by photovoltaic drivers. Each half H-bridge configuration in a channel is controlled through a dual P-N channel MOSFET. Consequently, a single actuation channel can be controlled using only two control pins, preventing potential short circuits. The same schematic can be replicated in a modular fashion to create multiple independent channels. A level shifter serves as the interface between the I/O pins of the microcontroller and the P-N channel MOSFETs.

\subsubsection*{Charge retention}
    During testing, we noticed the same charge retention effects recently described by Rumley et al. \cite{Rumley2022ChargeRetention}. As seen in Fig. \ref{Sfig:measurement}B strain drops slowly during actuation and stabilizes after a few seconds. When the voltage signal is turned off, the actuator remains in a non-zero actuation state due to internal fields caused by charge accumulation. Reversing the polarity mostly mitigates this effect \cite{Rumley2022ChargeRetention}. We saw a correlation between the contamination of the dielectric oil and the strength of charge retention. For example, dielectric oil exposed to humidity or dust performed considerably worse. We tried using dielectric oils with higher moisture tolerance, such as Midel 7131 but encountered problems such as chemical incompatibility with other actuator materials.

\subsection*{Text S1. Mean value theorem for effective dielectric constant}
    Kellaris er al. \cite{kellaris2019analytical} compute the total free electrical energy of the system $U_e$ as
    \begin{equation}\label{eq2}
        U_e = - \frac{1}{2}C(\alpha)V^2 = -\frac{w l_e(\alpha)}{4t}\epsilon_0 \epsilon_rV^2
    \end{equation}
    Eq.~\ref{eq2} assumes a constant permittivity, which applies to linear dielectrics. For nonlinear dielectrics, this approach can be extended with the energy density of a dielectric. The energy density of a dielectric material is equal to the integral of the discharge D-E curve~\cite{Chu2006}
    \begin{equation}
    u_e = \int EdD
    \end{equation}
    \begin{equation}
    u_e = \int_0^{D}E(\hat{D})d\hat{D} = \int_0^{D}\frac{1}{\epsilon_0\epsilon_r(\hat{D})}\hat{D}d\hat{D}
    \end{equation}
    We use the mean value theorem for integrals to show the existence of $\Tilde{D}$:
    \begin{align}
    \exists\Tilde{D}\in[0,\epsilon(D)E]:u_e=\frac{1}{\epsilon_0\epsilon_r(\Tilde{D})}\int_0^{\epsilon_0\epsilon_r(D)E}\hat{D}d\hat{D}=\frac{1}{\epsilon_0\epsilon_r(\Tilde{D})}\big[\frac{1}{2}\epsilon_0^2\epsilon_r(D)^2E^2\big]\\=\frac{1}{2}\epsilon_0\frac{\epsilon_r(D)^2}{\epsilon_r(\Tilde{D})}E^2
    \label{eq:ue}
    \end{align}
    with
    \begin{equation}
    \frac{\epsilon_r(D)^2}{\epsilon_r(\Tilde{D})}=\epsilon_{eff}
    \label{eq:eps}
    \end{equation}
    and
    \begin{equation}
    \hat{D}=\epsilon_r(\hat{D})\hat{E}\epsilon_0; \Tilde{D}=\epsilon_r(\Tilde{D})\Tilde{E}\epsilon_0; D=\epsilon_r(D)E\epsilon_0
    \end{equation}
    This is a non-constructive proof and we resort to measuring $\epsilon_{eff}$. Using Eq.~\ref{eq:ue} and Eq.~\ref{eq:eps} the total free electrical energy of the system is
    \begin{equation}
    U_e = \int u_edV = wl_e(\alpha)tu_e = wl_e(\alpha)t \frac{1}{2}\epsilon_0\epsilon_{eff}E^2=\frac{wl_e(\alpha)}{4t}\epsilon_0\epsilon_{eff}V^2
    \end{equation}
    We obtain the force equation by following the same steps as shown by Kellaris et al. \cite{kellaris2019analytical}. We add the free electrical energy to the free mechanical energy, minimize the result with respect to $\alpha$ and solve for the force, which gives us
    \begin{equation}\label{eq4}
        F = wt \frac{cos(\alpha)}{1-cos(\alpha)} \epsilon_0 \epsilon_{eff}E^2
    \end{equation}

\FloatBarrier
\clearpage
    
\FloatBarrier
\clearpage
    
    \begin{figure}[h]
        \centering
        \includegraphics[width = 0.9\textwidth]{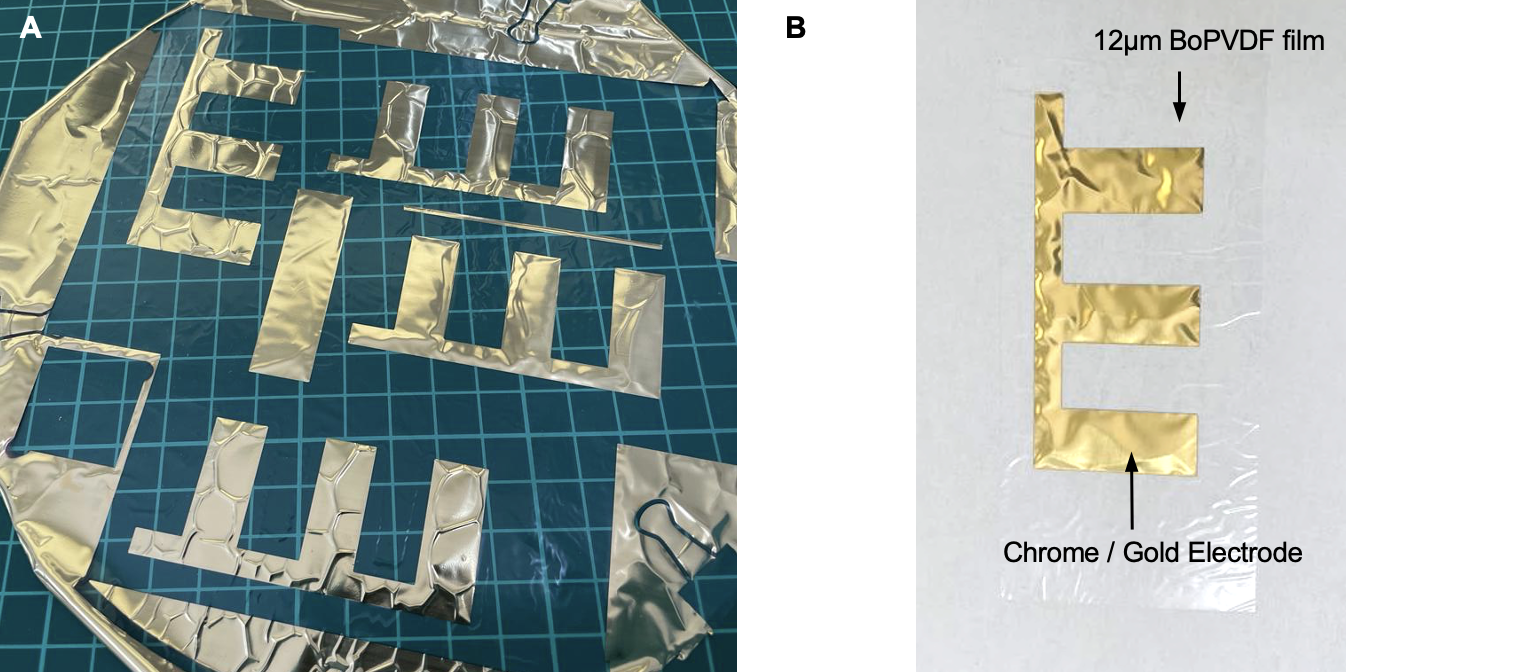}
        \caption{\textbf{Manufacturing of BoPVDF structural shell and gold electrode with additive manufacturing.} (\textbf{A})~A thin film of BoPVDF coated with chrome and gold with vapor deposition. An acrylic mask was used to define the electrode shapes. (\textbf{B})~Single side of an actuator after cutting, ready to be coated with a dielectric.}
        \label{figS:BoPVDFElectrodeManufacturing}
    \end{figure}
    
\FloatBarrier
\clearpage
    
\FloatBarrier
\clearpage

\FloatBarrier
\clearpage

\FloatBarrier
\clearpage

\begin{figure}[h]
\centering
\includegraphics[width = \textwidth]{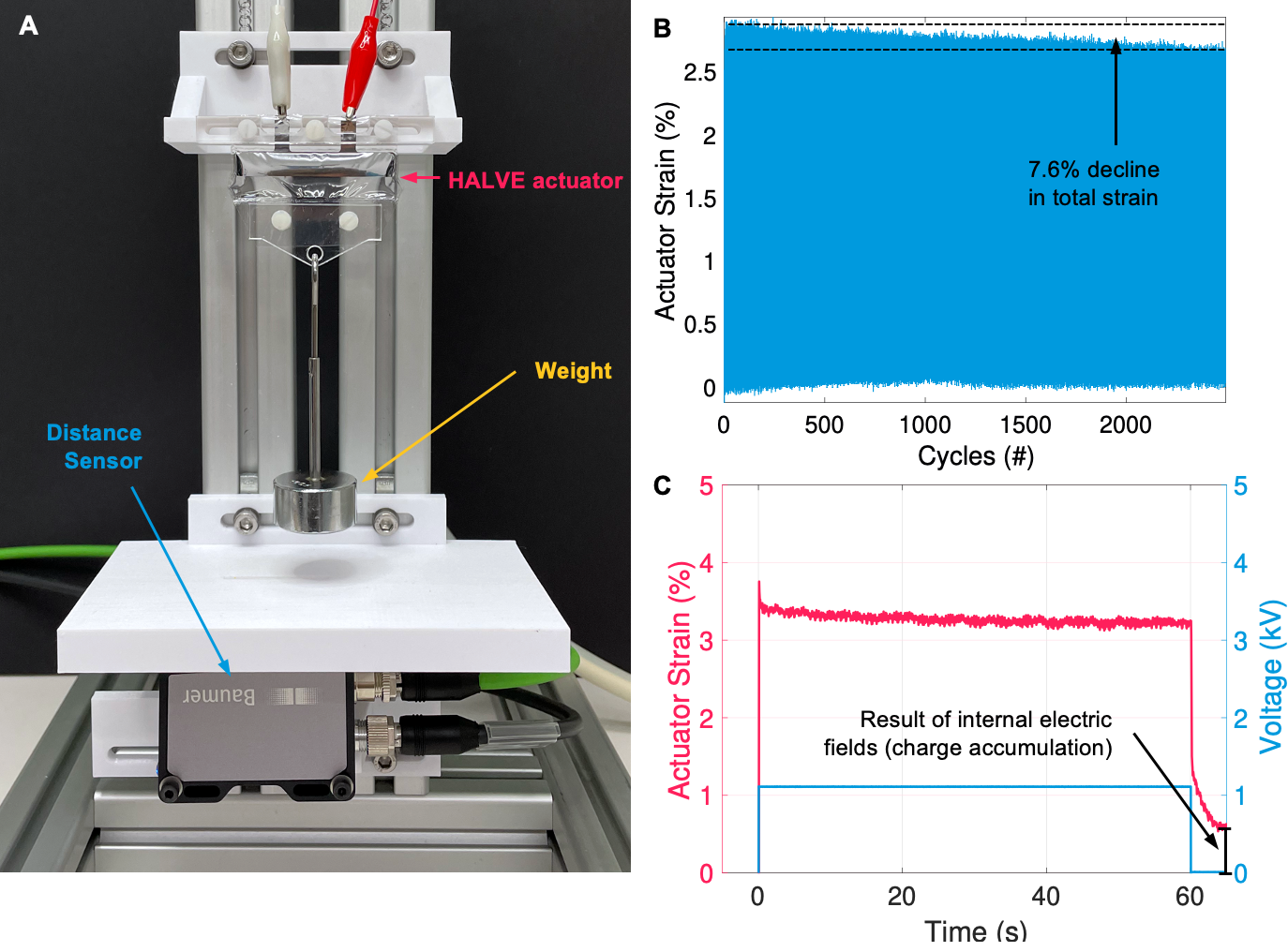}
\caption{\textbf{Characterization setup, \hl{HALVE actuator} durability test, and actuation characteristics.}
(\textbf{A})~Characterization setup used for actuator characterization. The setup consists of an actuator mounting location and a laser distance sensor.
(\textbf{B})~\hl{HALVE actuator} durability test, showing strain data of a \hl{HALVE actuator} (Aluminum/BoPET) lifting \qty{200}{\gram} with a bipolar signal of \qty{800}{\volt} at \qty{1}{\hertz}. A \qty{7.6}{\percent} decline in total actuation strain can be seen after 2500 actuation cycles .
(\textbf{C})~\hl{HALVE actuator} strain response to a \qty{1100}{\volt} step input over 60 seconds.}
\label{Sfig:measurement}
\end{figure}

\FloatBarrier
\clearpage

    \begin{figure}[h]
        \centering
        \includegraphics[width = \textwidth]{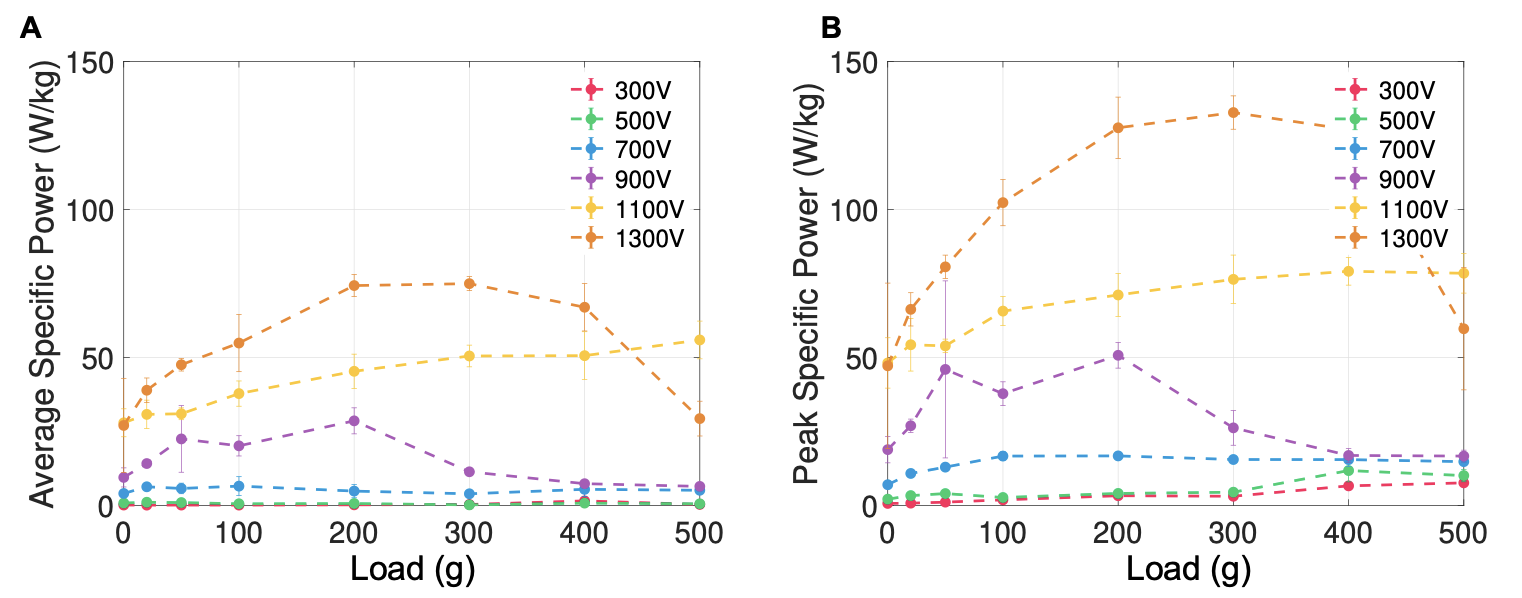}
        \caption{\textbf{Average and peak specific power of \hl{HALVE actuator} devices} (\textbf{A})~Average specific power plotted against applied load for an actuation voltage range of \qtyrange{300}{1300}{\volt}.} (\textbf{B})~Peak specific power plotted against applied load for an actuation voltage range of \qtyrange{300}{1300}{\volt}.
        \label{figS:specificPowerGraph}
    \end{figure}

\FloatBarrier
\clearpage

\FloatBarrier
\clearpage

\FloatBarrier
\clearpage

\begin{figure}[h]
    \centering
    \includegraphics[width = 0.9\textwidth]{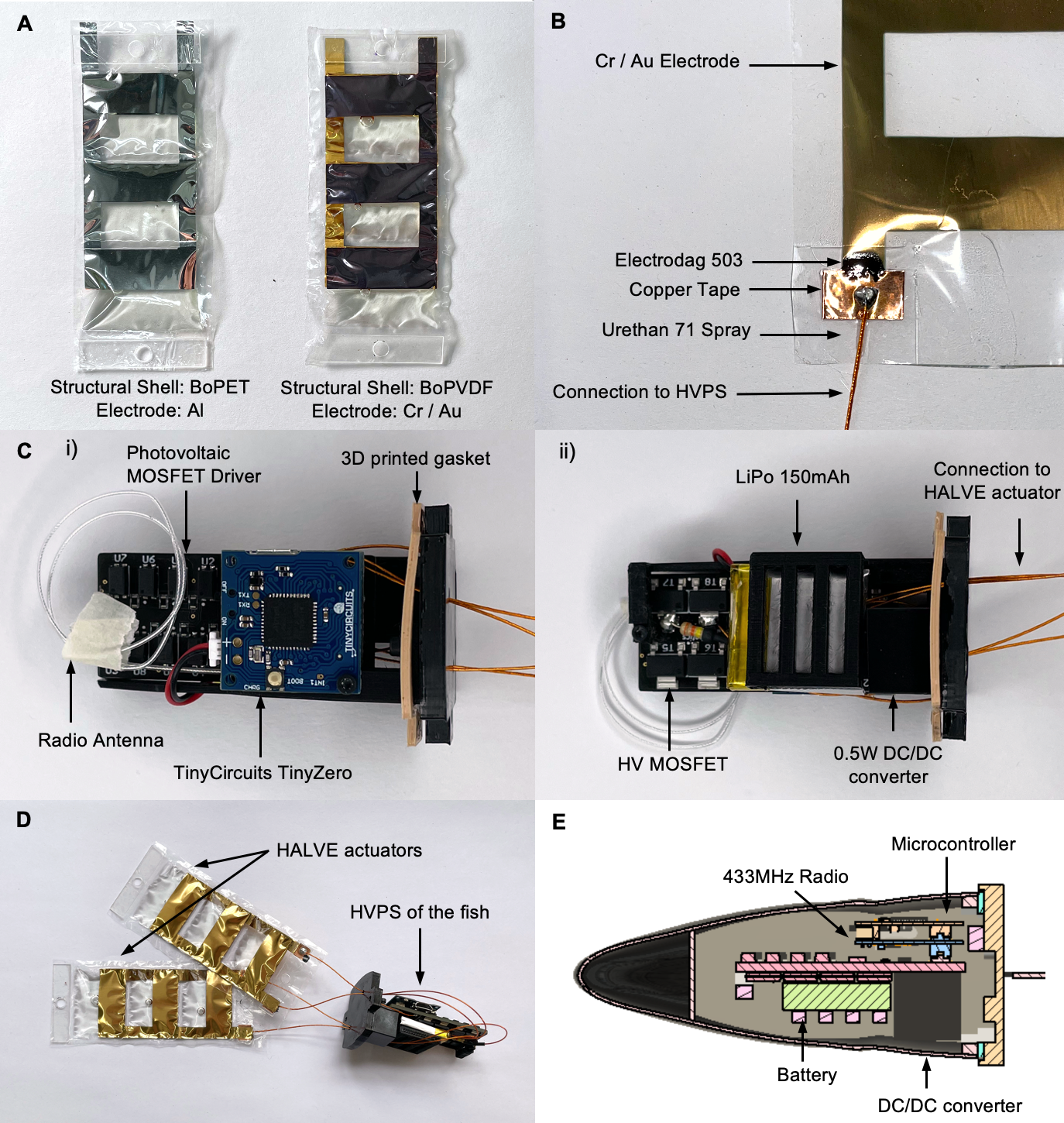}
    \caption{\textbf{\hl{HALVE actuators} and power supply system of the untethered robotic fish.}
(\textbf{A})~\hl{HALVE actuators} produced for the robotic fish. The BoPET and Aluminium actuators on the left were not used for the fish, as the pouch seal was deemed too unreliable for this application.
(\textbf{B})~ Two-channel power supply used to power the robotic fish, connected to a supporting structure which is part of the fish head.
(\textbf{C})~ Water-proof electrical connection between a \hl{HALVE actuator} electrode and a copper wire, achieved by soldering a wire to conductive copper tape, which was then adhered to the \hl{HALVE actuator}. Conductive ink (Electrodag 503) was then applied to the connection, which was then waterproofed by applying Urethane 71 spray.
(\textbf{D})~ Bottom view of the two-channel power supply.
(\textbf{E})~ \hl{HALVE actuators} and power supply assembly used in the robotic fish.
(\textbf{F})~ Cross-sectional diagram of the robotic fish head.}
    \label{figS:ActuatorFish}
\end{figure}

\FloatBarrier
\clearpage

\begin{figure}[h]
    \centering
    \includegraphics[width = \textwidth]{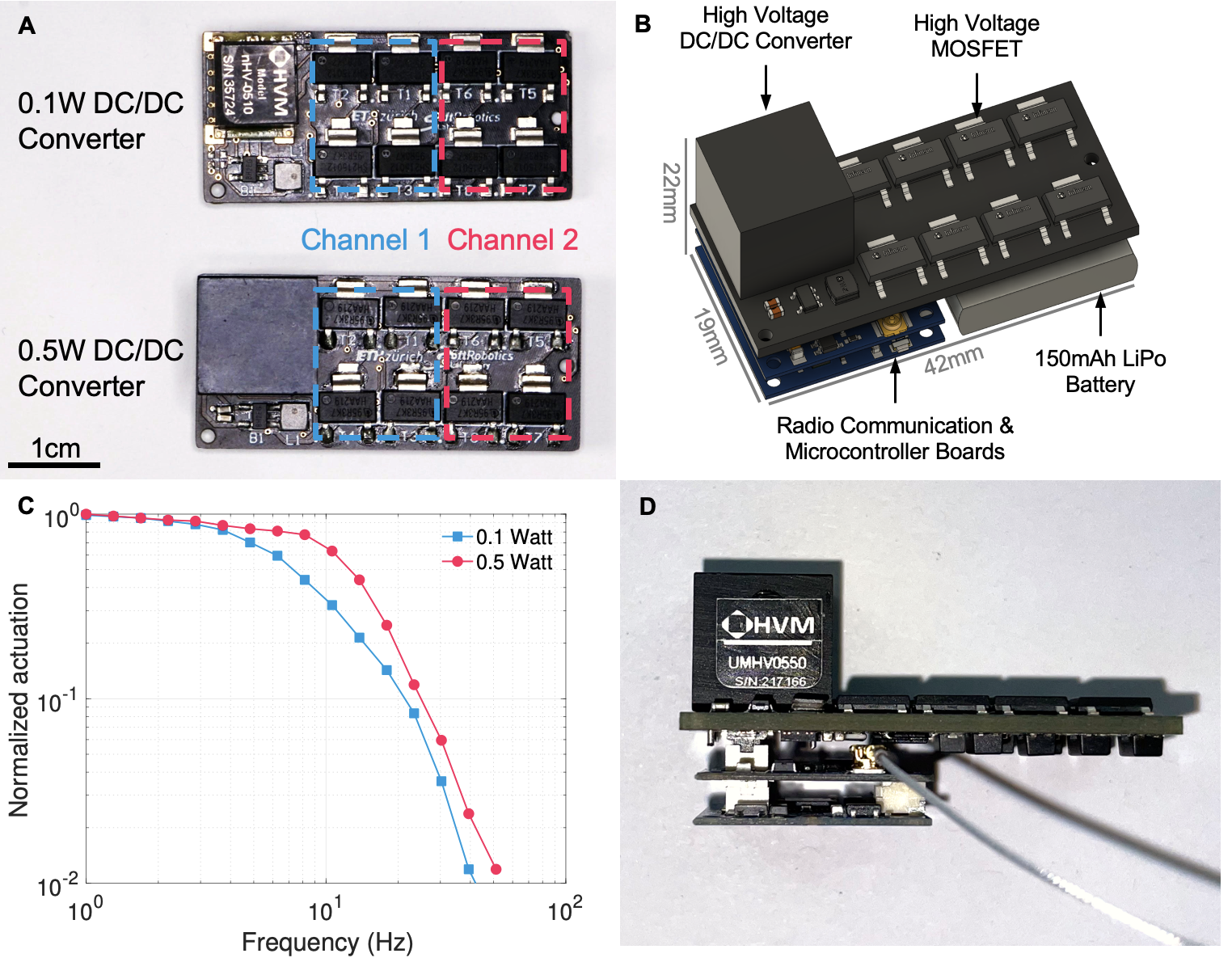}
    \caption{\textbf{Design of the compact high-voltage power supply (HVPS).} (\textbf{A}) The top view of two HVPS designs with different power capacities (\qty{0.1}{\watt} and \qty{0.5}{\watt}) is presented, where each power supply features two bipolar channels. (\textbf{B}) The computer-aided design (CAD) representation of the 0.5 W HVPS encompasses the battery, processor board, and radio module for communication purposes. (\textbf{C}) The frequency response of a \hl{HALVE actuator} when subjected to a step input excitation signal is illustrated. (\textbf{D}) The front view of the \qty{0.5}{\watt} HVPS displays the processor and radio module boards.}
    \label{figS:electronics}
\end{figure}

\FloatBarrier
\clearpage

\FloatBarrier
\clearpage

\begin{figure}[h]
    \centering
    \includegraphics[width = \textwidth]{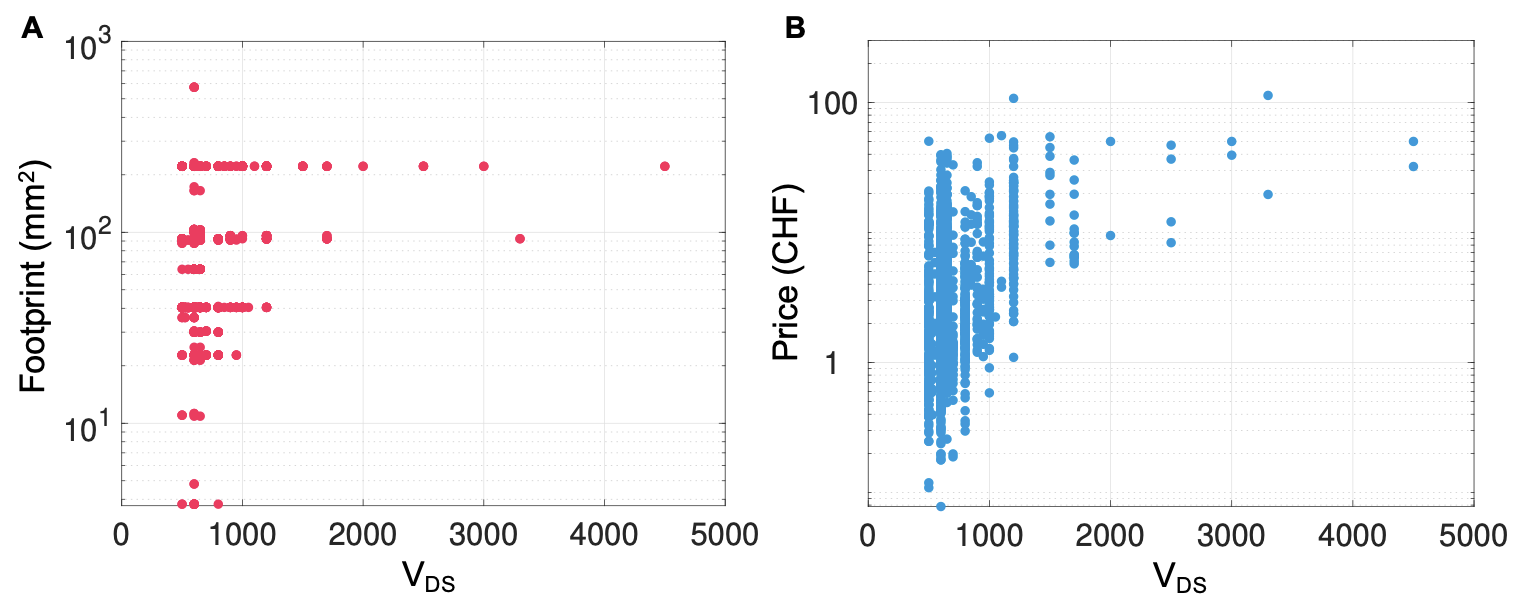}
    \caption{\textbf{Footprint and cost of high voltage MOSFETs.} (\textbf{A})~1745 datapoints were collected from the online catalog "Digi-Key electronics" in May 2022. All MOSFETs on the website were filtered for N-channel, surface-mounted, and $V_{DS}>499V$. Footprint size of a power supply given its maximum output voltage. (\textbf{B})~Component price given its maximum output voltage.}
    \label{figS:storyStuff}
\end{figure}

\FloatBarrier
\clearpage

\begin{table}[h]
\centering
\begin{tabular}{lll}
\toprule
            & Custom HV optocoupler (OZ100SG)   & MOSFET (Infineon IPN95R3K7P7m) \\ \midrule
Control     & \qty{265}{\milli\watt}~\cite{mitchell2022pocket}            & \qty{13.2}{\milli\watt}~\cite{MOSFET_datasheet}    \\
on leakage  & \qty{1000}{\milli\watt}~\cite{Opto_datasheet}           & \qty{1.85}{\milli\watt}~\cite{MOSFET_datasheet}    \\
off leakage & \qty{2.5}{\milli\watt}~\cite{Opto_datasheet}            & \qty{0.95}{\milli\watt}~\cite{MOSFET_datasheet}    \\
Max voltage & \qty{10}{\kilo\volt}~\cite{Opto_datasheet}              & \qty{0.95}{\kilo\volt}~\cite{MOSFET_datasheet}      \\
Price       & \$55                              & \$0.70    \\
Volume      & \qty{240}{\cubic\milli\meter}~\cite{mitchell2022pocket}     & \qty{75}{\cubic\milli\meter}~\cite{MOSFET_datasheet}     \\
Weight      & \qty{1.2}{\gram}~\cite{mitchell2022pocket}                  & \qty{0.11}{\gram}~\cite{MOSFET_datasheet}    \\\bottomrule
\end{tabular}
\caption{\textbf{Comparison between high voltage switching components.} Comparison of a state-of-the-art high voltage switching optocoupler used by Mitchell et al.~\cite{mitchell2022pocket} to the MOSFET device used in this work.}
\label{tabS:switchingComponent}
\end{table}

\FloatBarrier
\clearpage

\begin{table}[h]
\begin{tabular}{lp{5cm}ll}
\toprule
\textbf{Designator} & \textbf{Description}                             & \textbf{Part Number} & \textbf{Weight (g)} \\ \midrule
U1             & 12 Bit Digital to Analog Converter       & MCP4921-E/SN       & 0.016  \\ 
U2, U3, U4, U5 & Photovoltaic MOSFET Driver               & APV2111V           & 0.043  \\
U6                  & Boost Switching Regulator IC   Positive Fixed 5V & RP402N501F-TR-FE     & 0.013                                       \\
U7             & Level Shifter                            & NTS0103GU10,115    & 0.052  \\
U8 (0.1W)      & High Voltage DC-DC Converter 1000V 100mW & NHV0510            & 1.581  \\
U8 (0.5W)      & High Voltage DC-DC Converter 1000V 500mW & UMHV0510           & 4.124  \\
U9             & Level Shifter                            & NTS0104BQ,115      & 0.016 \\
Q1, Q2         & MOSFET Array N and P-Channel             & IRF7509TRPBF       & 0.026  \\
Q3, Q4, Q5, Q6 & Transistor MOSFET N-Channel              & IPN95R3K7P7        & 0.115 \\
C1, C3, C5     & 0.1UF Capacitor                          & GRM188R72A104KA35D & 0.006  \\
C2, C4, C6     & 10UF Capacitor                           & GRM188R60J106ME47D & 0.006  \\
R1, R2, R5, R6 & Resistor 500 OHM                         & RT0603BRC07500RL   & 0.002  \\
R3, R4         & Resistor 10K OHM                         & RMCF0603JT10K0     & 0.002  \\
L1             & 2.2UH Inductor                           & VLS3012HBX-2R2M    & 0.049  \\\bottomrule
\end{tabular}
\caption{\textbf{Power supply component list.} List of all components used for the high voltage power supply.}
\label{tabS:listOfComponents}
\end{table}

\clearpage %Flush all the figures 

\end{document}